%% file: main.tex
\documentclass{article}

\usepackage{microtype}
\usepackage{graphicx}
\usepackage{caption}
\usepackage{subcaption}
\usepackage{subcaption} 
\usepackage{booktabs} % for professional tables
\usepackage{wrapfig}
\usepackage{pifont}
\usepackage{placeins}
\usepackage{soul}
\usepackage{enumitem}
\usepackage{bbm}
\usepackage{color,soul}
\usepackage{hyperref}

\usepackage[accepted]{icml2025}

\usepackage{amsmath}
\usepackage{amssymb}
\usepackage{mathtools}
\usepackage{amsthm}
\usepackage{float}
\input{math_commands.tex}

\newcommand{\cirone}
{\text{\ding{172}}}
\newcommand{\cirtwo}
{\text{\ding{173}}}
\newcommand{\cirthree}
{\text{\ding{174}}}
\newcommand{\cirfour}
{\text{\ding{175}}}

\usepackage[capitalize,noabbrev]{cleveref}

\ifodd 0

\newcommand{\com}[1]{\textbf{\color{blue}([KY]: #1)}}
\newcommand{\CommentGabe}[1]{\textcolor[rgb]{0,0,1}{[Gabe comment: #1]}}

\newcommand{\CommentWong}[1]{\textcolor[rgb]{1,0,0}{[Wong comment: #1]}}

\else

\newcommand{\com}[1]{}
\newcommand{\CommentWong}[1]{}

\newcommand{\CommentGabe}[1]{}

\renewcommand{\st}[1]{}
\fi

\theoremstyle{plain}
\newtheorem{theorem}{Theorem}[section]

\newtheorem{lemma}[theorem]{Lemma}
\newtheorem{corollary}[theorem]{Corollary}
\theoremstyle{definition}
\newtheorem{definition}[theorem]{Definition}
\newtheorem{assumption}[theorem]{Assumption}
\theoremstyle{remark}

\usepackage[textsize=tiny]{todonotes}

\icmltitlerunning{NTK-DFL: Enhancing DFL in Heterogeneous Settings via NTK}

\begin{document}

\twocolumn[
\icmltitle{NTK-DFL: Enhancing Decentralized Federated Learning in Heterogeneous Settings via Neural Tangent Kernel}



\icmlsetsymbol{equal}{*}

\begin{icmlauthorlist}
\icmlauthor{Gabriel Thompson}{ece}
\icmlauthor{Kai Yue}{ece}
\icmlauthor{Chau-Wai Wong}{ece,sci}
\icmlauthor{Huaiyu Dai}{ece}
\end{icmlauthorlist}

\icmlaffiliation{ece}{Electrical and Computer Engineering, NC State University}
\icmlaffiliation{sci}{Secure Computing Institute, NC State University, Raleigh, USA}

\icmlcorrespondingauthor{Chau-Wai Wong}{chauwai.wong@ncsu.edu}

\icmlkeywords{Machine Learning, ICML, Federated Learning, Distributed Training, Decentralized Federated Learning, Neural Tangent Kernel}

\vskip 0.3in
]



\printAffiliationsAndNotice{}  % leave blank if no need to mention equal contribution

\begin{abstract}
Decentralized federated learning (DFL) is a collaborative machine learning framework for training a model across participants without a central server or raw data exchange. DFL faces challenges due to statistical heterogeneity, as participants often possess data of different distributions reflecting local environments and user behaviors. Recent work has shown that the neural tangent kernel (NTK) approach, when applied to federated learning in a centralized framework, can lead to improved performance. We propose an approach leveraging the NTK to train client models in the decentralized setting, while introducing a synergy between NTK-based evolution and model averaging. This synergy exploits inter-client model deviation and improves both accuracy and convergence in heterogeneous settings. Empirical results demonstrate that our approach consistently achieves higher accuracy than baselines in highly heterogeneous settings, where other approaches often underperform. Additionally, it reaches target performance in 4.6 times fewer communication rounds. We validate our approach across multiple datasets, network topologies, and heterogeneity settings to ensure robustness and generalization. Source code for NTK-DFL is available at \href{https://github.com/Gabe-Thomp/ntk-dfl}{https://github.com/Gabe-Thomp/ntk-dfl}.

\end{abstract}

\section{Introduction}
\label{submission}

Federated learning (FL) is a machine learning paradigm in which multiple clients train a global model without the explicit communication of training data. In most FL scenarios, clients communicate with a central server that performs model aggregation. In the popular federated averaging~(FedAvg) algorithm~\citep{fedavg}, clients perform multiple rounds of stochastic gradient descent~(SGD) on their own local data, then send this new weight vector to a central server for aggregation. As FL gains popularity in both theoretical studies and real-world applications, numerous improvements have been made to address challenges, including communication efficiency, heterogeneous data distributions, and security concerns \citep{comm_efficient, li2020federated, deep_leakage}. 
To handle the performance degradation caused by data heterogeneity, many works have proposed mitigation strategies for FedAvg~\citep{karimireddy2020scaffold, li2020federated}.
Notably, some researchers have introduced the neural tangent kernel~(NTK), replacing the commonly-used SGD in order to improve the model convergence~\citep{yu2022tct, yue2022neuraltangentkernelempowered}.

Despite these advancements, the centralized nature of traditional FL schemes introduces the possibility for client data leakage, computational bottlenecks at the server, and high communication bandwidth demand \citep{fed_open_problems}. Decentralized federated learning (DFL) has been proposed as a solution to these issues \citep{DFL_survey}. In DFL, clients may communicate with each other along an undirected graph, where each node represents a client and each edge represents a communication channel between clients. 
While DFL addresses some of the issues inherent to centralized FL, both frameworks grapple with the challenge of statistical heterogeneity across clients.
Although mixing data on a central server could readily resolve this issue, transmitting raw, private training data from clients introduces privacy concerns, making FL and DFL approaches good candidates to address this challenge~\citep{dfl_survey_2}.
This paper focuses on the following research question: How can we design a DFL approach that effectively addresses statistical heterogeneity? 

We propose a 
%DFL 
method that exploits the NTK to evolve weights. We denote this paradigm NTK-DFL. Our approach combines the advantages of NTK-based optimization with the decentralized structure of DFL. The NTK-DFL weight evolution scheme makes use of the communication of client Jacobians, allowing for more expressive updates than traditional weight vector transmissions and improving performance under heterogeneity. Complementing this NTK-based evolution, we utilize a model averaging step that exploits inter-client model deviation, creating a global model with much better generalization than any local model. {We demonstrate that NTK-DFL maintains high performance even under aggressive compression measures. Through reconstruction attack studies, we also analyze how this compression affects data privacy.}
The contributions of this paper are threefold. 
\begin{enumerate}[left=0pt]
    \item The proposed NTK-DFL method achieves convergence with 4.6 times fewer communication rounds than existing approaches in heterogeneous settings. To the best of our knowledge, this is the first work leveraging NTK-based weight evolution for decentralized federated training.
    % \footnote{The implementation for NTK-DFL is available at \href{https://github.com/Gabe-Thomp/ntk-dfl}{https://github.com/Gabe-Thomp/ntk-dfl}.} 
    \item The effective synergy between NTK-based evolution and DFL demonstrates superior resilience to data heterogeneity with model averaging. 
    \item The NTK-DFL aggregated model 
    exhibits robust performance across various network topologies, datasets, data distributions, {and compression measures}. This performance is further supported by theoretical bounds demonstrating improved convergence rates.
    %of the average weight.
    % achieves at least 10\% higher accuracy than the average accuracy of individual client models. This aggregated model 
\end{enumerate}

\section{Related Work}

\textbf{Federated Learning (FL)}\quad FL was introduced by \citet{fedavg} as a machine learning approach that enables training a model on distributed datasets without sharing raw data. It attempts to address key issues such as data privacy, training on decentralized data, and data compliance for more heavily regulated data (e.g., medical imaging) \citep{fl_survey_zhang}. Despite its advantages, the centralized topology of FL introduces several challenges. These include potential privacy risks at the central server, scalability issues due to computational bottlenecks, and high communication overhead from frequent model updates between clients and the server \citep{fl_survey_mothukuri}.

\textbf{Decentralized Federated Learning (DFL)}\quad DFL aims to eliminate the need for a central server by connecting clients in a fully decentralized topology. 
\citet{dfedavg} adapted the FedAvg approach of multiple local SGD iterations to the decentralized setting. 
% leveraging momentum to improve model convergence and weight quantization to reduce total communication cost. 
Our NTK-DFL method may be viewed as building on this foundation, using the neural tangent kernel for more effective weight updates.
\citet{dispfl} proposed a method of DFL where each client possesses their own sparse mask personalized to their specific data distribution. 
\citet{dfedsam} employed the sharpness-aware minimization optimizer to reduce the inconsistency of local models, whereas we tackle this issue through per-round averaging and final model aggregation. DFL approaches can aim to train one global model, such as the case of many hospitals training a model for tumor classification with local, confidential images \citep{tumor_example}. They may also aim to train a personalized model for each client to perform better on the local data distribution. For example, different groups of smartphone users may use different emojis and would benefit from a personalized model \citep{pfl_survey}. Our method focuses on training a high-performing global model that generalizes well across all clients, offering improved convergence and resilience to data heterogeneity compared to existing DFL methods.

\textbf{Neural Tangent Kernel (NTK)}\quad NTK has primarily been used for the analysis of neural networks \citep{ntk_survey}, though it has recently seen use in the training of neural networks for FL~\citep{yue2022neuraltangentkernelempowered}. Introduced by \citet{ntk_jacot}, it shows that the evolution of an infinitely wide neural network converges to a kernelized model. This approach has enabled the analytical study of models that are well approximated by this infinite width limit \citep{ntk_linearity}. NTK has also been extended to other model types, such as the recurrent neural network \citep{ntk_rnn} and convolutional neural network \citep{ntk_cnn}. We instead use the linearized model of the NTK approximation as a tool for weight evolution.
Some studies have explored the integration of NTKs with FL. 
For instance, \citet{huang2021fl} applied the NTK analysis framework to study the convergence properties of FedAvg, while \citet{yu2022tct} extended NTK applications beyond theoretical analysis by training a convex neural network. Moreover, \citet{yue2022neuraltangentkernelempowered} replaced traditional SGD-based optimization with NTK-based evolution in a federated setting, where clients transmit Jacobian matrices to a central server that performs weight updates using NTK.

\section{Proposed Method: NTK-Based Decentralized Federated Learning}

\subsection{Problem Statement}
\label{section:problem statement}
We begin with a brief overview of centralized FL. The goal of centralized FL is to train a global model $\vw$ across $M$ clients with their private, local data  
$\mathcal{D}_i = \{(\vx_{i,j},\vy_{i,j})\}_{j=1}^{N_i}$, where $N_i$ is the number of training examples of the $i$th client. FL algorithms aim to
numerically solve the sample-wise optimization problem of $\operatorname*{min}_\vw F(\vw)$, where
$F(\vw) = \frac{1}{M} \sum_{i=1}^{M} N_i F_i(\vw)$ and 
$F_i(\vw) = \frac{1}{N_i}\sum_{j=1}^{N_i} \mathcal{\ell}(\vw; \vx_{i,j}, \vy_{i,j})$.

In the decentralized setting, an omnipresent global weight~$\vw$ is not available to clients in each communication round. Rather, each client possesses their own model $\vw_i$ that is trained in the update process. Following related DFL work~\citep{dfedsam, dfedavg}, we seek a global model $\vw$ that benefits from the heterogeneous data stored locally across clients and generalizes better than any individual client model $\vw_i$. 
A global or aggregated model may take the form \({\vw = \frac{1}{N}\sum_{i=1}^M N_i \vw_i}\), where ${N=\sum_{i=1}^M N_i}$.

\textbf{Notation}\quad 
Formally, we have a set of clients {$\mathcal{C} = \{1,\dots,i,\ldots,{M}\}$.} Each client is initialized with its weight $\vw_i^{(0)}\in\R^d$, 
where %$i$ denotes a specific client and 
$d$ is the size of the parameter vector and the superscript in $\vw_i^{(0)}$ denotes the initial communication round. Model training is done in a series of communication rounds denoted $k \in \{1, 2, ..., K\}$. Let the graph at round $k$ be $\mathcal{G}^{(k)} = (\mathcal{C}, E^{(k)})$, where $E^{(k)}$ is the set of edges representing connections between clients.
Furthermore, the neighborhood of client $i$ at round $k$ is denoted ${\mathcal{N}_i^{(k)} = \{j\mid (i, j) \in E^{(k)}\}}$. This graph is specified before each communication round and can take an arbitrary form. 
\begin{figure}
    \centering
    \includegraphics[width=0.50\textwidth]{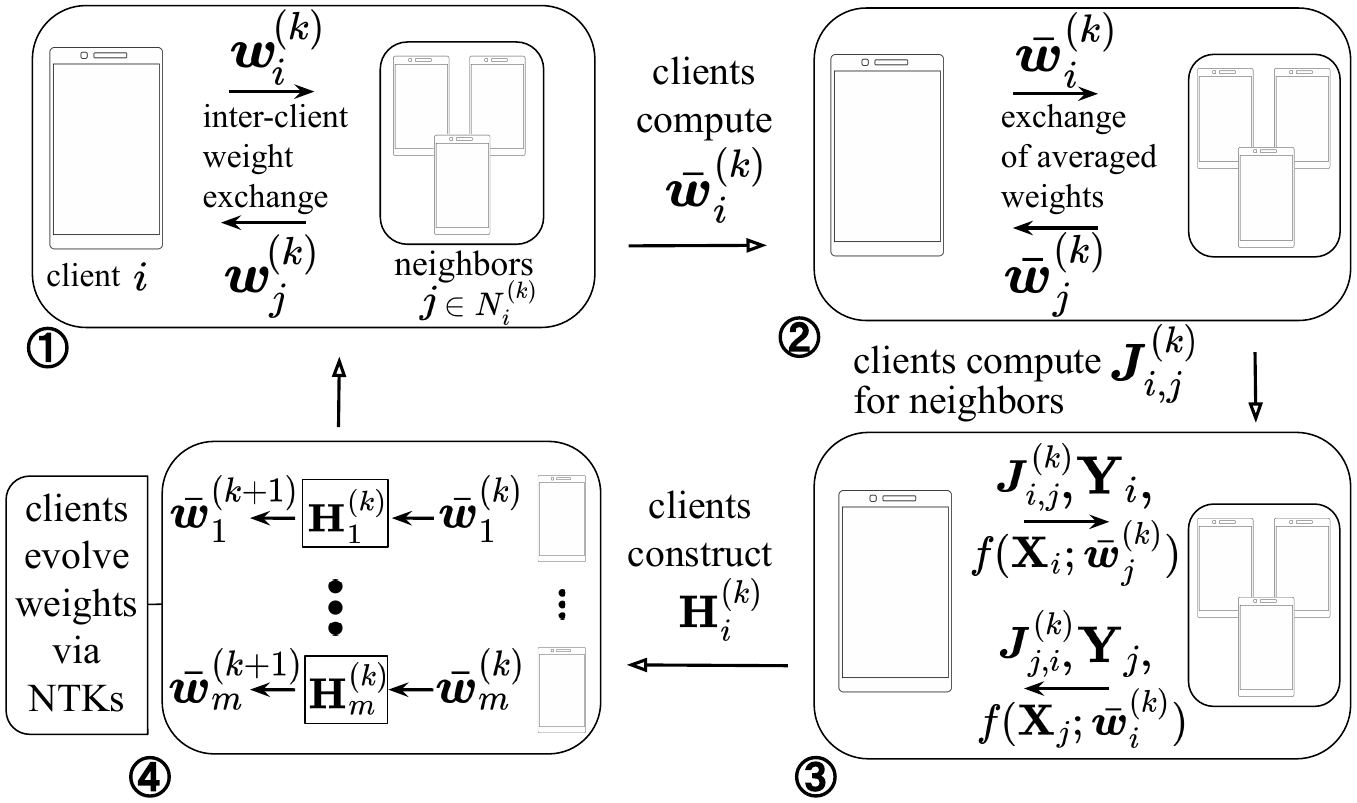}
    \caption{NTK-DFL process: \cirone\,Clients exchange weights, \cirtwo\,Average weights with neighbors, \cirthree\,Compute and exchange Jacobians, labels, and function evaluations, \cirfour\,Construct local NTK and evolve weights [Eq.~(\ref{eq:weights_dntk})]. This decentralized approach enables direct client collaboration and NTK-driven model evolution without a central server.
    \vspace{-4.9mm}
    }
    \label{fig:ntk-dfl-diagram}
\end{figure}

\subsection{Proposed NTK-DFL}
\label{section:Communication Protocol}
Figure \ref{fig:ntk-dfl-diagram} illustrates the proposed NTK-DFL method. We describe its key components below.
%, including the communication protocol and weight evolution process.

\textbf{Per-round Parameter Averaging}\quad At the beginning of each communication round $k$, each client $i$ both sends and receives weights---client $i$ sends its model $\vw_i^{(k)}$ to all neighbors ${j\in \mathcal{N}_i^{(k)}}$ and receives $\vw_j^{(k)}$ from {all} neighbors. Each client then aggregates its own weights with its neighbors' weights to form a new averaged weight as follows: 
\begin{equation}
\bar{\vw}_i^{(k)} = \tfrac{1}{{N_i + \sum_{j\in \mathcal{N}^{(k)}_i}N_j}} \big( {N_i\vw_i^{(k)} + \sum_{j \in \mathcal{N}_i^{(k)}} N_j\vw_j^{(k)} }\big).
\end{equation}
The client must then send this aggregated weight $\bar{\vw}_i^{(k)}$ back to all neighbors. This step enables each client to
construct a local NTK, comprised of inner products of Jacobians from both neighboring clients and its own Jacobian. See Algorithm~\ref{alg:param-avg} in Appendix~\ref{appendix:algorithms} for details.

\textbf{Local Jacobian Computation}\quad At this point, each client possesses its own aggregated weight $\bar{\vw}_i^{(k)}$ as well as a set of aggregated weights $\bar{\vw}_j^{(k)}$ from each of their neighbors ${j \in \mathcal{N}_i^{(k)}}$. The $i$th client
% use these weights $\bar{\vw}_j^{(k)}$ and local data $\mathbf{X}_i$ to 
computes the Jacobian of $f(\mathbf{X}_i;\bar{\vw}_j^{(k)})$ with respect to the neighboring model parameters {$\bar{\vw_j}^{(k)}$} using its local data $\mathbf{X}_i$. We denote this neighbor-specific Jacobian as 
\begin{equation}
\mJ^{(k)}_{i,j} \equiv [\nabla_{{\vw}}f(\mathbf{X}_i;\bar{\vw}_j^{(k)})]^\top.
\end{equation}
Each client sends every neighbor their respective Jacobian $\mJ^{(k)}_{i,j}$, true label $\mathbf{Y}_i$, and function evaluation $f(\mathbf{X}_i;\bar{\vw}_j^{(k)})$. Note the order of the indices in the Jacobian{:} the $i$th client {sends} $\mJ_{i,j}^{(k)}$, an evaluation on the $i$th {client's} data and its {neighbors'} weights. In contrast, the $i$th client {receives} $\mJ_{j,i}^{(k)}$ from each of its neighbors, an evaluation on its {neighbors'} data and the $i$th {client's} weights. 
Algorithm \ref{alg:jacobian-computation-sending} in Appendix~\ref{appendix:algorithms} describes this process.

\textbf{Weight Evolution}\quad
After all inter-client communication is completed, the clients begin the weight evolution phase of the round (see Algorithm \ref{alg:weight-evolution} in Appendix \ref{appendix:algorithms}). Here, all clients act in parallel as computational nodes. Each client possesses their own Jacobian tensor $\mJ_{i,i}^{(k)}$ as well as their neighboring Jacobian tensors $\mJ_{j,i}^{(k)}$ for each ${j \in \mathcal{N}_i^{(k)}}$.

We denote the tensor of all Jacobian matrices possessed by the $i$th client at round $k$ as $\bm{\mathcal{J}}_i^{(k)}$, which is composed of matrices from the set $\{\mJ_{i,i}^{(k)}\} \cup \{\mJ_{j,i}^{(k)}\ |\ {j \in \mathcal{N}_i^{(k)}}\}$ stacked along the third dimension. We denote the matrix of true labels and function evaluations stacked in the same manner as $\bm{\mathcal{Y}}_i$ and $\vf(\bm{\mathcal{X}}_i)$, respectively. 
% Here, $i$ denotes a client index $i\in \mathcal{C}$. Each tensor $\bm{\mathcal{J}}_i$, $\bm{\mathcal{Y}}_i$, and $\vf(\bm{\mathcal{X}}_i)$ is a stacked representation of the data from each client and its neighbors. 
Explicitly, we have $\bm{\mathcal{J}}_i^{(k)}\in \R^{\tilde{N_i}\times d_2 \times d}$, $\bm{\mathcal{Y}}_i^{(k)}\in \R^{\tilde{N_i}\times d_2}$, and $\vf(\bm{\mathcal{X}}_i)\in \R^{\tilde{N_i}\times d_2}$. 
Additionally, $\tilde{N_i} = N_i + \sum_{j \in \mathcal{N}_i^{(k)}} N_j$ represents the total number of data points between client $i$ and its neighbors, and $d_2$ is the output dimension.

% \medskip
From here, each client performs the following operations to evolve its weights. First, compute the local NTK $\mathbf{H}_i^{(k)}$ from the Jacobian tensor $\bm{\mathcal{J}}_i^{(k)}$ using the definition of NTK: 
\begin{equation}
[\mathbf{H}^{(k)}_{i}]_{m,n}=\frac{1}{d_2}\langle \bm{\mathcal{J}}_i^{(k)}(\vx_{m}),\bm{\mathcal{J}}_i^{(k)}(\vx_{n})\rangle_\text{F}.
\end{equation}
Each element of the NTK is a pairwise Frobenius inner product between Jacobian matrices, where the indices $m$ and $n$ correspond to the $m$th and $n$th data points, respectively. Second, using $\mathbf{H}_i^{(k)}$, the client evolves their weights as follows (see Appendix \ref{appendix:C} for more details):
% \begin{equation}
%    \label{eq:analytical_dntk}
% \vf^{(k,t)}(\bm{\mathcal{X}}_i) = \bigl( \mathbf{I} \! - \! e^{-\frac{\eta t}{\tilde{N}_i} \mathbf{H}_i^{(k)}} \bigr) \bm{\mathcal{Y}}_i^{(k)} 
% \! + \! e^{-\frac{\eta t}{\tilde{N}_i} \mathbf{H}_i^{(k)}} \vf^{(k)}(\bm{\mathcal{X}}_i),
% \end{equation}
\begin{equation}
\label{eq:analytical_dntk}
\begin{split}
\vf^{(k,t)}(\bm{\mathcal{X}}_i) 
&= \bigl( \mathbf{I} \! - \! e^{-\frac{\eta t}{\tilde{N}_i} \mathbf{H}_i^{(k)}} \bigr) \bm{\mathcal{Y}}_i^{(k)} \\
&\quad + \, e^{-\frac{\eta t}{\tilde{N}_i} \mathbf{H}_i^{(k)}} \vf^{(k)}(\bm{\mathcal{X}}_i).
\end{split}
\end{equation}
\newpage
We unroll gradient steps to find the weight $\vw_i^{(k,t)}$ as follows:\begin{subequations}
\begin{align}
   \vw_i^{(k,t)}&=\sum_{j=1}^{d_2}(\bm{\mathcal{J}}^{(k)}_{i,:j:})^\top \mathbf{R}_{i,:j}^{(k,t)}+\bar{\vw}_i^{(k)},
   \label{eq:weights_dntk}\\
   \mathbf{R}_{i,:j}^{(k,t)}&\equiv\frac{\eta}{\tilde{N_i} d_2}\sum_{u=0}^{t-1}[\bm{\mathcal{Y}}_i^{(k)}-\vf^{(k,u)}(\bm{\mathcal{X}}_i)].
   \label{eq:weights_dntk2}
\end{align}
\end{subequations}
Third, the client selects the weight $\vw_i^{(k,t)}$ for a timestep~$t$ with the lowest loss according to the evolved residual $\vf^{(k,t)}(\bm{\mathcal{X}}_i)-\bm{\mathcal{Y}}_i$. This is used as the new weight $\vw_i^{(k+1,0)}$ for the next communication round.

\textbf{Final Model Averaging} \label{final_model_averaging}\quad
Throughout the paper, we study the convergence of the final averaged model $\bar{\vw}^{(k)} = \frac{1}{M}\sum_{i=1}^M \vw_i^{(k)}$. 
In the decentralized setting, clients would average all $K$ client models after all training is completed. This may be done through a fully-connected topology, sequential averaging on a ring topology, or in a secure, centralized manner. The average may also be carried out over a subset of clients. In practice, we observe that the aggregated model~$\bar{\vw}^{(k)}$ generally performs better than any individual client model~${\vw}_i^{(k)}$. We study the impact of the order of client averaging on model performance with a {client selection algorithm} {and show the results in} Figure \ref{fig:Selection Algorithm}. Each client that opts in to model averaging contributes a portion of its data to a global validation set before training begins. Our client selection algorithm selects clients in the order of their accuracy on the validation set. We will demonstrate that in the practical setting, with a proper selection of clients, not all nodes must opt into final model averaging in order for the aggregated model to benefit from improved convergence. We note a difference between \textit{model consensus}, often discussed in the DFL literature \citep{dfl_consensus_1, dfl_consensus_2}, and the proposed final model averaging approach. Model consensus refers to the eventual convergence of all client models to a single, unified model over numerous communication rounds. In contrast, our approach implements final model averaging as a distinct step performed after the completion of the training process.

Lastly, while communication overhead and memory efficiency are not the primary focus of this paper, we briefly note a technique to address potential memory constraints in NTK-DFL implementations. For scenarios involving dense networks or large datasets, we introduce Jacobian batching. This approach allows clients to process their local datasets in smaller batches, reducing memory complexity from $O(N_i d_2 d)$ to $O(N_i d_2 d/m_1)$, where $m_1$ is the number of batches. 
% Clients compute and transmit Jacobians for each batch separately, evolving their weights multiple times per communication round. This complexity reduction allows clients to connect in a denser network for the same memory cost. 
{We also study communication efficiency, where NTK-DFL is resilient to compression measures such as top-$k$ sparsification and random projections. This enables significant reductions in communication costs without significantly compromising convergence.} 
% Notably, we are able to reduce the communication cost by a factor of about 30 by using these methods (see Figure \ref{fig:comm_compare_bits}) in Appendix \ref{appendix:mitigation}.
% A thorough discussion of network and communication overhead can be found in Appendix \ref{appendix:mitigation}.
A thorough discussion of overhead can be found in Appendices~\ref{appendix:tradeoffs} and~\ref{appendix:mitigation}. We also provide results on a reconstruction attack performed using client Jacobians in Appendix~\ref{appendix:recon-attack}.

\section{{Theoretical Analysis}}
In this section, we derive a convergence bound for NTK-DFL by analyzing the behavior of the average client weight $\bar{\vw}^{(k)} = \frac{1}{M}\sum_{i=1}^M \vw_i^{(k)}$, similar to \citet{dfedavg, dfedsam}. {Unlike DFedAvg, the NTK-DFL bound includes a key additional dependence on the number of local iterations $T$ in its main term. 
% The bound highlights that NTK-DFL’s uniquely large $T$ values are a key factor driving its improved convergence.
The bound captures the unique advantage that NTK-DFL gains by using much larger $T$ values compared to other methods, improving convergence.
% The bound effectively captures the positive effect that a large $T$, which is unique to NTK-DFL, has on convergence.
}
The analysis also captures key factors such as the relationship between the spectral gap and model convergence, impact of data heterogeneity, and NTK approximation error. %We introduce the necessary definitions and assumptions below.
% \newpage
\begin{definition} (DFL objective). We are interested in minimizing the global loss $\mathcal{L}(\vw)$ across all clients, defined as
\begin{equation}
   \mathcal{L}(\!\vw\!) \!=\! \sum_{i=1}^{M} \! \tfrac{N_i}{N_{\text{total}}} \! \mathcal{L}_i(\!\vw\!),\ \mathcal{L}_i(\vw) \!=\! \tfrac{1}{N_i} \! \sum_{j=1}^{N_i} \! \ell(\!\vw; {\vx}_{i,j},\!\mathbf{\vy}_{i,j}\!),
\end{equation}
where \(\ell(\cdot)\) is a sample-wise loss applied to client data.
% We note that most of these assumptions are mild and commonly used in the DFL literature \citep{d-psgd, dfedavg, dfedsam}.
% \CommentGabe{NOTE: These assumptions are similar to other DFL works. I need to make sure they are paraphrased well enough}
\end{definition} 
\begin{definition}
\label{def:mixing}
(The gossip/mixing matrix). [Definition 1, \citep{dfedavg}] The matrix \( \mathbf{M} = [m_{i,j}] \in [0,1]^{m \times m} \) is assumed to satisfy the following properties: (i) (Graph) If \( i \neq j \) and \( (i,j) \notin \mathcal{V} \), then \( m_{i,j} = 0 \); otherwise, \( m_{i,j} > 0 \). (ii) (Symmetry) The matrix is symmetric, i.e., \( \mathbf{M} = \mathbf{M}^{\top} \). (iii) (Null space property) The null space of \( \mathbf{I} - \mathbf{M} \) is spanned by the all-ones vector, \( \text{null}(\mathbf{I} - \mathbf{M}) = \text{span}\{\mathbbm{1}\} \). (iv)~(Spectral property) The matrix satisfies \( \mathbf{I} \succeq \mathbf{M} \succ -\mathbf{I} \). The eigenvalues of \( \mathbf{M} \) satisfy \( 1 = |\lambda_1(\mathbf{M})| > |\lambda_2(\mathbf{M})| \geq \dots \geq |\lambda_m(\mathbf{M})| \), and the spectral gap is defined as \( (1 - \lambda)^2 \in (0,1] \), where \( \lambda := \max\{|\lambda_2(\mathbf{M})|, |\lambda_m(\mathbf{M})|\} \).
\end{definition}
\begin{assumption}
    (Standard DFL assumptions).
    We assume that $\nabla \mathcal{L}_i$ is Lipschitz continuous. We bound the variance of client gradients with~$\sigma_g^2$ and the client gradient norm with~$B$. We note that these assumptions are relatively mild and common in the DFL literature. Additional details can be found in Appendix \ref{appendix:math}.
\end{assumption}
\begin{assumption}
\label{assum:ntk-error}
(Approximation error of NTK gradient). For any client \( i \in \{1,2,\dots,M\} \), the difference between the NTK gradient and the true gradient of the loss is bounded by \( \delta_{\text{NTK}} \), i.e.,
\begin{equation}
\left\|\nabla \mathcal{L}_i^{\text{NTK}}(\vw) - \nabla \mathcal{L}_i(\vw) \right\|^2 \leq \delta_{\text{NTK}}^2,
\end{equation}
for all \( \vw \in \mathbb{R}^d \). We note that an explicit formulation of $\delta_{\text{NTK}}$ is possible given assumptions of a simplified model \citep{huang2021fl, yue2022neuraltangentkernelempowered} (see Lemma \ref{lem:delta_ntk_bound} in Appendix \ref{appendix:math}). In our analysis, we treat this as a constant in order to remain model-agnostic.
\end{assumption}
\begin{theorem}
\label{thm:main} Consider the average weight over $M$ clients $\bar{\vw}^{(k)}\coloneqq \frac{1}{M  }\sum_{i=1}^M \vw_i^{(k)}$ for round $k\in\{1,\ldots,K\}$ and suppose the previously stated assumptions hold. Assuming that the learning rate satisfies $0<\eta\leq\frac{1}{8LT}$, we have
\begin{multline} 
\min_{1 \leq k \leq K} \left\| \nabla \mathcal{L}\left( \bar{\vw}^{(k)} \right) \right\|^2  \leq 
\frac{2\bigr[\mathcal{L}( \bar{\vw}^{(1)} ) - \mathcal{L}^*\bigl]}{K\gamma(T,\eta)}\\
+\alpha(\eta,T,\sigma_g,\delta_{\text{NTK}})+\beta(\eta, T,\sigma_g,\delta_{\text{NTK}},\lambda),   
\end{multline}
where the constants are defined as 
% \begin{multline}
%     \gamma(T,\eta) \coloneqq 
%     \left( \eta T - 32\eta^2T^2L^3\eta^2(\eta T L + 1) \right),\ 
%     \\\alpha(\eta,T,\sigma_g,\delta_{\text{NTK}}) \coloneqq 
%     \frac{1}{\gamma(T,\eta)} \Bigl[\bigl( \eta L^2 T + L \bigr) \\
%         \cdot \bigl( 16 \eta^2 T^2 (\delta_{\text{NTK}}^2 + \sigma_g^2) + 2\eta T \delta_{\text{NTK}} B \bigr)\Bigr],\\
%     \beta(\eta, T, \sigma_g, \delta_{\text{NTK}}, \lambda) \coloneqq
%     32\eta^2T^2L^3\eta^2(\eta T L + 1) \\\cdot 
%     \frac{\bigl[ 16 T^2 (\delta_{\text{NTK}}^2 + \sigma_g^2) + 16 T^2 B^2 \bigr]}{(1 - \lambda)^2\gamma(T,\eta)}.
% \end{multline}
% \begin{multline}
%     \gamma(T,\eta) \coloneqq 
%     \left( \eta T - 32\eta^2T^2L^3\eta^2(\eta T L + 1) \right),\ 
%     \\\alpha(\eta,T,\sigma_g,\delta_{\text{NTK}}) \coloneqq 
%     \frac{1}{\gamma(T,\eta)} \Bigl[\bigl( \eta L^2 T + L \bigr) \\
%         \cdot \bigl( 16 \eta^2 T^2 (\delta_{\text{NTK}}^2 + \sigma_g^2) + 2\eta T \delta_{\text{NTK}} B \bigr)\Bigr],\\
%     \beta(\eta, T, \sigma_g, \delta_{\text{NTK}}, \lambda) \coloneqq
%     32\eta^2T^2L^3\eta^2(\eta T L + 1) \\\cdot 
%     \frac{\bigl[\delta_{\text{NTK}}^2 + \sigma_g^2 + B^2 \bigr]}{(1 - \lambda)^2\gamma(T,\eta)}.
% \end{multline}
% \end{theorem}
\begin{subequations}
\begin{align}
\gamma(T,\eta) 
&\coloneqq 
    \eta T - 32\eta^2T^2L(\eta T L + 1),\ 
    \\
\alpha(\eta,T,\sigma_g,\delta_{\text{NTK}}) &\coloneqq 
    \tfrac{1}{\gamma(T,\eta)} 2 \eta T \big[ L ( \eta L T + 1) \notag{} \\
    &\ \ \ \cdot 8 \eta T (\delta_{\text{NTK}}^2 \! + \! \sigma_g^2) + \delta_{\text{NTK}} B \big],\\
\beta(\eta, T, \sigma_g, \delta_{\text{NTK}}, \lambda) &\coloneqq
    \tfrac{1}{\gamma(T,\eta)} 512\eta^4T^4L^3 (\eta T L + 1)\notag{}\\
    &\ \ \ \cdot 
    (1 - \lambda)^{-2} (\delta_{\text{NTK}}^2 \! + \! \sigma_g^2 \! + \! B^2).
\end{align}
\end{subequations}
\end{theorem}
\begin{corollary}
Let the learning rate satisfy $O(1/L\sqrt{KT})$. Using similar assumptions as Theorem \ref{thm:main} in Appendix \ref{appendix:math}, we have the following convergence rate for NTK-DFL:
\begin{multline}
    \min_{1 \leq k \leq K} \left\| \nabla \mathcal{L}\left( \bar{\vw}^{(k)} \right) \right\|^2  \leq O\biggl(\frac{\mathcal{L}( \bar{\vw}^{(1)} ) - \mathcal{L}^*}{\sqrt{KT}}\\ + \frac{\sqrt{T}(\delta_{\text{NTK}}B+\delta_{\text{NTK}}^2+\sigma_g^2)}{\sqrt{K}}+
    \frac{T(\delta_{\text{NTK}}^2+\sigma_g^2)}{K}\\
    + \frac{T(\delta_{\text{NTK}}^2+\sigma_g^2+B^2)}{(1-\lambda)^2K}\biggr)+O(\delta_{\text{NTK}}).
\end{multline}
\end{corollary}
From the bound above, we can see that the main convergence term can be improved when we increase the number of local iterations, $T$. However, as expected, we cannot arbitrarily increase $T$ without inducing a greater error in the terms $\delta_{\text{NTK}}$, $B$, and $\sigma_g$. Compared to bounds in \citet{dfedavg}, 
$O\Big(\frac{1}{\sqrt{K}}+\frac{\sigma_g^2}{\sqrt{K}}+\frac{\sigma_g^2+B^2}{{(1-\lambda)^2K^{3/2}}}\Big)$, the first term to the right of the inequality highlights improved convergence derived from the ability to select a value of $T$ that is 1 to 2 orders of magnitude larger than that of other DFL methods. Intuitively, the bound tightens with an increasing spectral gap $(1-\lambda)^2$, which is associated with mixing speed. We note that there is an irreducible convergence floor from the NTK approximation term $\delta_\text{NTK}$. Obtaining an explicit bound on this term can be done through specific assumptions about the loss function and the model itself. For instance, as detailed in Lemma \ref{lem:delta_ntk_bound}, assumptions about the smoothness of $f_i$ and a bounded per-client residual norm lead to $\delta_{\text{NTK}} = O\big(K^{-1}T (\sigma_g^2+B^2)\big)$, thus becoming reducible.
% for large~$K$.

\input{tex_figures/fig_rounds_fmnist_selected}
\input{tex_figures/fig_neighbor_hetero}

\section{Experiments}
\subsection{Experimental Setup}

\textbf{Datasets and Model Specifications}\quad Following \citet{yue2022neuraltangentkernelempowered}, we experiment on three datasets: Fashion-MNIST \citep{fashionmnist}, FEMNIST \citep{leaf}, and MNIST \citep{mnist}. Each dataset contains $C=10$ output classes. For Fashion-MNIST and MNIST, data heterogeneity has been introduced in the form of non-IID partitions created by the symmetric Dirichlet distribution \citep{dirichlet}. For each client, a vector $\vq_i\sim \text{Dir}(\alpha)$ is sampled, where $\vq_i\in \R^{C}$ is confined to the $(C-1)$-standard simplex such that $\sum_{j=1}^{C}q_{ij}=1$. This assigns a probability distribution over labels to each client, creating heterogeneity in the form of label-skewness. For smaller values of $\alpha$, a client possesses a distribution concentrated in fewer classes. We test over a range of $\alpha$ values in order to simulate different degrees of heterogeneity. In FEMNIST, data is split into shards based on the writer of each digit, introducing heterogeneity in the form of feature-skewness. For the model, we use a two-layer multilayer perceptron with a hidden width of 100 neurons for all trials.

\textbf{Network Topologies}\quad  A sparse, time-variant $\kappa$-regular graph with $\kappa = 5$ {was} used as the standard topology for experimentation, where for each communication round $k$, a new random graph $\mathcal{G}^{(k)}$ with the same parameter $\kappa$ is created. Various values of $\kappa$ {were} tested to observe the effect of network density on model convergence. We also experimented with various topologies to ensure robustness to different connection settings. We used a network of 300 clients throughout our experiments. 

\textbf{Baseline Methods}\quad We compare our approach to various state-of-the-art baselines in the DFL setting. These include D-PSGD \citep{d-psgd}, DFedAvg, DFedAvgM \citep{dfedavg}, DFedSAM \citep{dfedsam}, and DisPFL \citep{dispfl}. We also compare with the centralized baseline NTK-FL \citep{yue2022neuraltangentkernelempowered}. 
The upper bound NTK-FL would consist of a {client fraction} of 1.0 where the server constructs an NTK from all client data each round, which is infeasible due to memory constraints. Instead, we conducted a comparison following \citet{dispfl}, which we include in Appendix~\ref{appendix:baselines}. We also include details on baseline hyperparameters in Appendix~\ref{appendix:baseline-hyperparameters}.
% , equalizing the degree of the busiest node in DFL with that of the server in centralized FL. 
% \EditGabe{Additional details regarding baseline hyperparameters can be found in Appendix \ref{appendix:baseline-hyperparameters}}. 
% \EditGabe{with general details of baselines in Appendix \ref{appendix:baselines} and hyperparameters in Appendix \ref{appendix:baseline-hyperparameters}}.

\textbf{Performance Metrics}\quad We evaluate the performance of the various DFL approaches by studying the aggregate model accuracy on a global, holdout test set. This ensures that we are measuring the generalization of the aggregate model from individual, heterogeneous local data to a more representative data sample. Our approach is in line with the goal of training a global model capable of improved generalization over any single local model (Section \ref{section:problem statement}), unlike personalized federated learning where the goal is to fine-tune a global model to each local dataset \citep{pfl_survey}. When evaluating the selection algorithm in Figure \ref{fig:Selection Algorithm}, we split the global test set in a 50:50 ratio of validation to test data. We use the validation data to sort the models based on their accuracy, and report the test accuracy in the figure.

\subsection{Experimental Results}
\textbf{Test Accuracy \& Convergence}\quad Our experiments demonstrate the superior convergence properties of NTK-DFL compared to baselines. 
Figure~\ref{fig:fashion_mnist_convergence} illustrates the convergence trajectories of NTK-DFL and other baselines on Fashion-MNIST.
We see that NTK-DFL convergence benefits are {enhanced} under increased heterogeneity. 
Under high heterogeneity with $\alpha = 0.1$, NTK-DFL establishes a 3--4\% accuracy lead over the best-performing baseline within just five communication rounds and maintains this advantage throughout the training process. Additionally presented are the number of communication rounds necessary for convergence to 85\% test accuracy, where NTK-DFL consistently outperforms all baselines. 
For the $\alpha=0.1$ setting, NTK-DFL achieves convergence in 4.6 times fewer communication rounds than DFedAvg, the next best performing baseline.
Figure \ref{fig:different_dataset_convergence} in Appendix \ref{appendix:B} demonstrates a similar convergence advantage for NTK-DFL on both feature-skewed FEMNIST and label-skewed MNIST datasets.

\textbf{Factor Analyses for NTK-DFL}\quad We evaluate NTK-DFL's performance over various factors, including the sparsity and heterogeneity levels, the choices of the topology, and weight initialization scheme.
Figure~\ref{fig:topology_compare} illustrates the test accuracy of NTK-DFL and other baselines as functions of the sparsity and heterogeneity levels, respectively. 
We observe a mild increase in convergence accuracy with decreasing sparsity. NTK-DFL experiences stable convergence across heterogeneity values $\alpha$ ranging from $0.1$ to $0.5$.
The left plot reveals that NTK-DFL consistently outperforms baselines by 2--3\% across all sparsity levels.
The right plot demonstrates NTK-DFL's resilience to data heterogeneity---while baseline methods' performance deteriorates with decreasing~$\alpha$, NTK-DFL maintains stable performance. 
In Figure~\ref{fig:all topology} of Appendix~\ref{appendix:B}, we evaluate NTK-DFL across a range of network topologies and find that it performs consistently well across different connection structures with the same sparsity level. Additionally, Figure~\ref{fig:dynamic} illustrates the impact of a dynamic network topology on NTK-DFL convergence. The dynamic topology accelerates convergence compared to the static topology, likely due to improved information flow among clients. Figure~\ref{fig:weight init} demonstrates the effect of weight initialization on NTK-DFL performance. While random per-client initialization slightly slows convergence compared to uniform initialization, NTK-DFL exhibits robustness to these initialization differences.

\input{tex_figures/fig_heterogeneity_compare}

\begin{figure}
    % \centering
    % \vspace{-14pt}
    \includegraphics[width=1\linewidth]{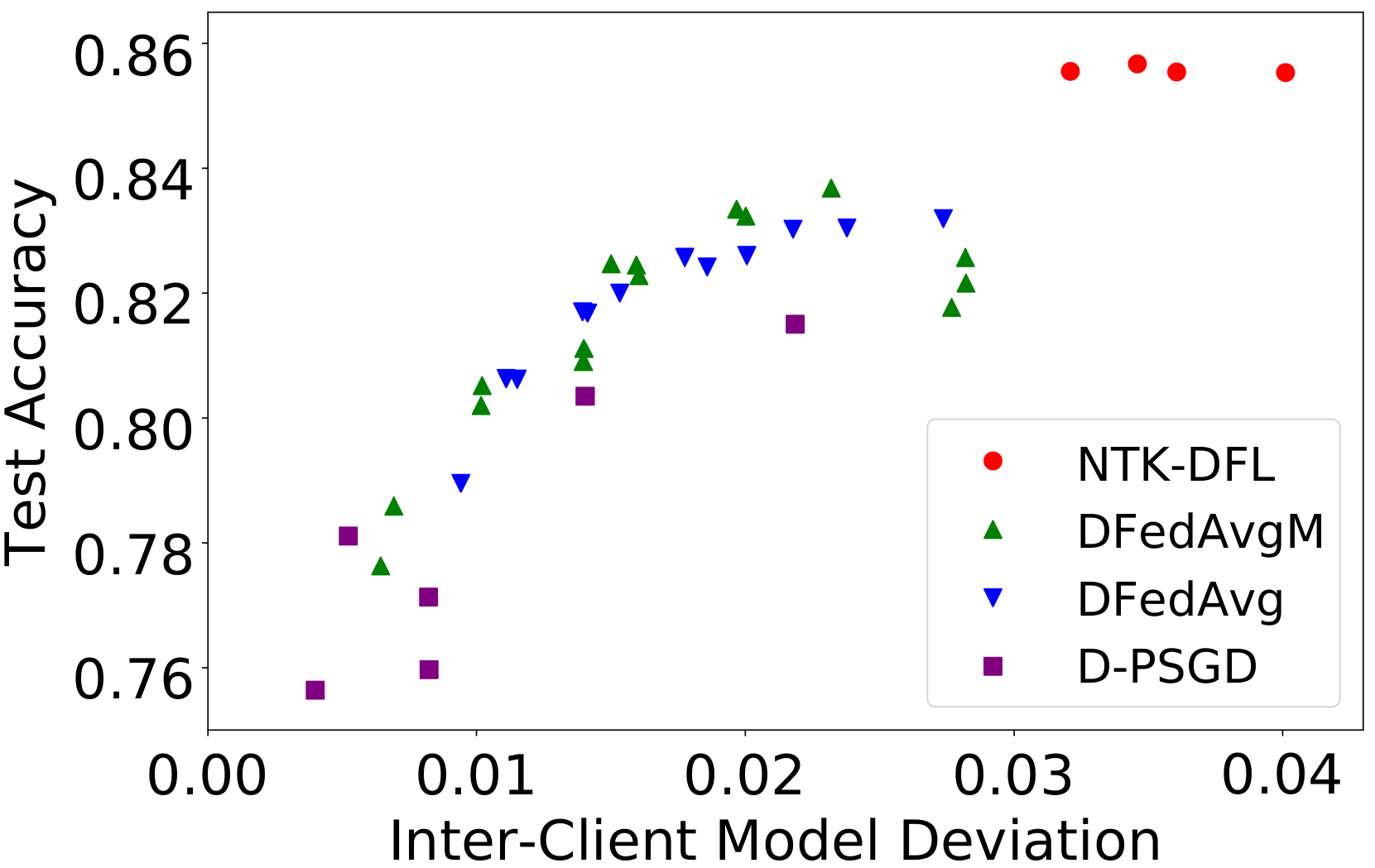}
    % \captionsetup{width=0.45\textwidth, margin={3pt,0pt}}
    \caption{Relationship between inter-client model deviation and final test accuracy {on Fashion-MNIST}. Each point represents a trial with distinct hyperparameters. The plot reveals a positive correlation between model deviation and accuracy, suggesting that higher deviation may benefit model averaging in DFL to a certain extent. Notably, the NTK-DFL approach demonstrates both higher accuracy and greater model deviation compared to other methods.}
    \label{fig:var_vs_acc}
\end{figure}
\textbf{Gains Due to Final Model Aggregation}
\quad Figure~\ref{fig:heterogeneity_compare} 
demonstrates the dramatic effect of final model aggregation on final test accuracy. 
Though the individual client models decrease in accuracy as the level of heterogeneity increases, the final aggregated model remains consistent across all levels of heterogeneity (as seen in Figures~\ref{fig:fashion_mnist_convergence} and~\ref{fig:topology_compare}). 
In the most heterogeneous setting $\alpha=0.1$ that we tested, the difference between the mean accuracy of each client and the aggregated model accuracy is nearly 10\%, as shown in Figure \ref{fig:heterogeneity_compare}. A similar phenomenon is observed in Figure~\ref{fig:acc_vs_sparsity_level} of Appendix~\ref{appendix:B} as the client topology becomes more sparse. For the same heterogeneity setting with a sparser topology of $\kappa=2$, the difference between these accuracies is nearly 15\%. Though the individual performance of local client models may suffer under extreme conditions, the inter-client model deviation (see detailed definition in  Appendix~\ref{subsec:inter-client-def}) created by such unfavorable settings is exploited by model averaging to recuperate much of that lost performance. 
\begin{figure*}[!t]
    \centering
    
    \begin{minipage}[t]{0.49\textwidth}
        \centering
        \includegraphics[width=1.035\linewidth]{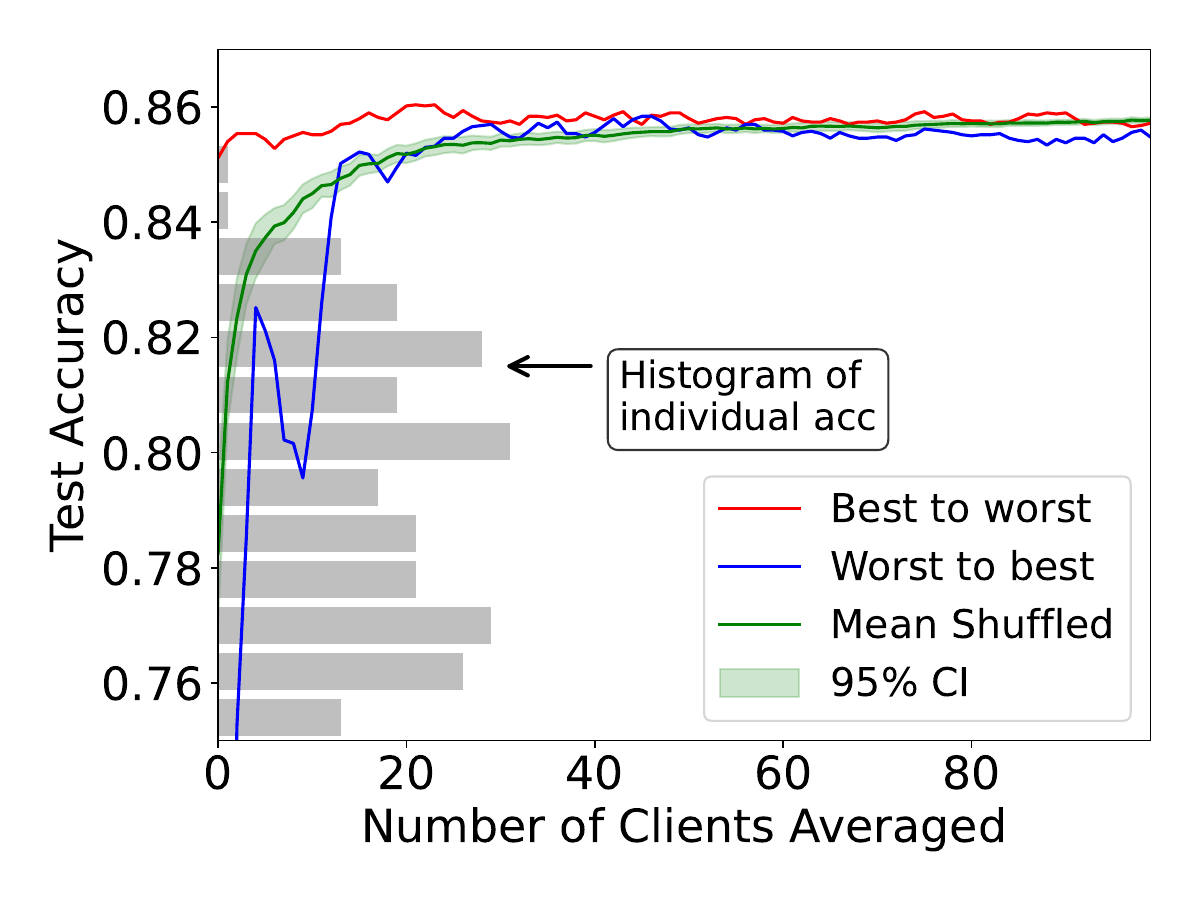}
        \captionsetup{width=0.45\textwidth, margin={10pt,0pt}}
        \caption{Final model test accuracy {on Fashion-MNIST} vs.~the number of clients averaged for a highly heterogeneous setting with $\alpha=0.1$.
         The histogram shows the distribution of individual client model accuracies. Three client selection criteria are tested: proposed high-to-low (red), random (green), and low-to-high (blue). 
         }
        \label{fig:Selection Algorithm}
    \end{minipage}
    \hfill
    \begin{minipage}[t]{0.49\textwidth}
        % \vspace{3pt}
        {\centering
        \includegraphics[width=1.04\linewidth]{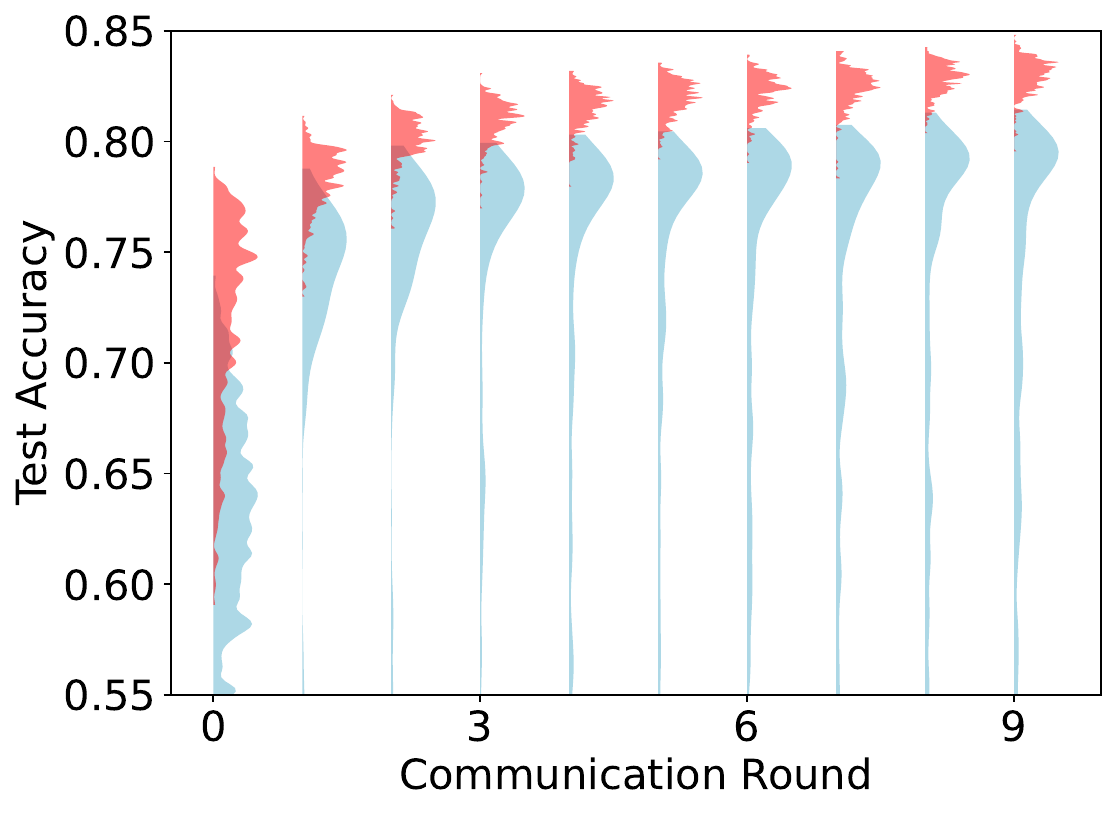}}
        % \captionsetup{width=0.5\textwidth, margin={0pt,0pt}}
        \caption{Distributions of {individual} client model accuracy vs. the communication round for Fashion-MNIST.
        The proposed scheme~(red) conducts per-round averaging among neighbors, whereas the ablated setup~(blue) does not. 
        Per-round averaging reduces the skewness of the model performance distribution.}
        \label{fig:Averaging ablation}
    \end{minipage}%
    
    \vspace{-5pt}
\end{figure*}
Figure~\ref{fig:var_vs_acc} suggests 
that inter-client model deviation enhances the performance of model averaging in DFL.
While extreme dissimilarity in model weights would likely result in poor performance of the averaged model, we observe that a moderate degree of deviation can be beneficial. For example, this is seen in~\citet{fedprox} with the tuning of the “proximal term”~$\mu$ that regulates this degree of client diversity in model updates. We posit that the NTK-based update steps generate a more advantageous level of deviation compared to baseline approaches, contributing to improved overall performance.

\textbf{Selection Algorithm}
\quad Figure~\ref{fig:Selection Algorithm} demonstrates the results of the selection algorithm for the final model aggregation. The ``best-to-worst'' selection algorithm is highly effective in a highly heterogeneous setting with $\alpha=0.1$.
It significantly outperforms a random averaging order and the lower-bound averaging order, which requires the fewest clients to be averaged to achieve the same level of accuracy. In practical deployments, final model aggregation has implications in a fully decentralized setting. For example, when training starts to converge, clients may need to connect in a denser topology to shortlist neighbors with higher validation accuracies for final model aggregation. %This approach could optimize the efficacy of the final aggregation step while maintaining the decentralized nature of the system.

\textbf{Per-round Averaging Ablation Study}\quad  In Figure~\ref{fig:Averaging ablation}, we perform an ablation study in which we remove the per-round parameter averaging that is a part of the NTK-DFL process. Here, clients forego the step of averaging their weight vectors with their neighbors during each communication round. Instead, clients compute Jacobians with respect to their original weight vector and send these to each of their neighbors (see Algorithm~\ref{alg:jacobian-computation-sending} in Appendix~\ref{appendix:algorithms}). A massive distribution shift can be seen in the figure, where the distribution in the ablated setting is clearly skewed into lower accuracies. In contrast, NTK-DFL with per-round averaging demonstrates a much tighter distribution around a higher mean accuracy, effectively eliminating the long tail of low-performing models. Per-round averaging in NTK-DFL serves as a stabilizing mechanism against local model drift, safeguarding clients against convergence to suboptimal solutions early in the training process.
In other words, client collaboration in the form of per-round averaging with neighbors ensures that no client lags behind in convergence. This is a particularly valuable feature in decentralized federated learning scenarios, where maintaining uniformity across a diverse set of clients with heterogeneous data is a major challenge \citep{DFL_survey}.

\textbf{Communication and Memory Overhead}\quad While overhead is not the primary focus of our work, we propose several strategies to allow for control over these factors.
To mitigate memory overhead, we experiment with Jacobian batching, where clients divide their data into batches each round and perform an NTK-DFL update per batch. As shown in Figure~\ref{fig:jacobian_batching_compare} of Appendix~\ref{appendix:mitigation}, test accuracy per communication round improves with increasing batch size.
To address communication overhead, we explore datapoint subsampling, where clients compute Jacobians with respect to only a fraction of their data each round. Figure~\ref{fig:sampling} of Appendix~\ref{appendix:mitigation} illustrates the expected accuracy drop from this approach---a natural trade-off for reduced communication.
Finally, we evaluate compression techniques to reduce NTK-DFL’s communication volume further. Figure~\ref{fig:communication_comparison} of Appendix~\ref{appendix:mitigation} compares the convergence of NTK-DFL with different compression methods, while Figure~\ref{fig:comm_compare_bits} presents convergence in terms of bits communicated. We leave readers to Appendix~\ref{appendix:mitigation} for detailed discussions.

\section{Conclusion and Future Work}
In this paper, we have introduced NTK-DFL, a novel approach to decentralized federated learning that leverages the neural tangent kernel to address the challenges of statistical heterogeneity in decentralized learning settings. {Our work extends NTK-based training to the decentralized setting,
% through novel studies in Jacobian batching and datapoint subsampling
while discovering a unique synergy between NTK evolution and decentralized model averaging that improves final model accuracy.} Our method combines the expressiveness of NTK-based weight evolution with a decentralized architecture, allowing for efficient, collaborative learning without a central server. We reduce the number of communication rounds needed for convergence, which may prove advantageous for high-latency settings or those with heavy encoding/decoding costs. 

There are promising unexplored directions for NTK-DFL. For instance, extending the algorithm to training models such as CNNs, ResNets \citep{resnet}, and transformers \citep{attention_is_all_you_need}, {possibly making use of new NTK methods suited for modern architectures} \citep{ntk_cnn, ntkresnet, ntk_transformer}. Additionally, future research could explore the application of NTK-DFL to cross-silo federated learning scenarios, particularly in domains such as healthcare, where data privacy concerns and regulatory requirements often necessitate decentralized approaches \citep{cross_silo}. Lastly, NTK-DFL may serve as a useful paradigm for transfer learning applications in scenarios where a single, centralized source of both compute and data is not available.


% \textbf{Do not} include acknowledgements in the initial version of
% the paper submitted for blind review.

% If a paper is accepted, the final camera-ready version can (and
% usually should) include acknowledgements.  Such acknowledgements
% should be placed at the end of the section, in an unnumbered section
% that does not count towards the paper page limit. Typically, this will 
% include thanks to reviewers who gave useful comments, to colleagues 
% who contributed to the ideas, and to funding agencies and corporate 
% sponsors that provided financial support.

% \newpage 
\section*{Impact Statement}
This paper presents work aimed at advancing the field of machine learning. Our research has several potential societal consequences, none of which we believe need to be specifically highlighted in this context.
%\CommentGabe{Add something here? The ICML guidelines say it is ok to leave it blank if there is nothing in particular we want to say. Also, it doesn't count towards the page limit.}


\section*{Acknowledgments}

This work was supported in part by the US National Science Foundation under grants SaTC-2340856 and ECCS-2203214, and the ECE Undergraduate Research Program and the Caldwell Fellows Program at the NC State University. The views expressed in this publication are those of the authors and do not necessarily reflect the views of the National Science Foundation. 

% \bibliography{example_paper}

%%%%%%%%%%%%%%%%%%%%%%%%%%%%%%%%%%%%%%%%%%%%%%%%%%%%%%%%%%%%%%%%%%%%%%%%%%%%%%%
%%%%%%%%%%%%%%%%%%%%%%%%%%%%%%%%%%%%%%%%%%%%%%%%%%%%%%%%%%%%%%%%%%%%%%%%%%%%%%%
% APPENDIX
%%%%%%%%%%%%%%%%%%%%%%%%%%%%%%%%%%%%%%%%%%%%%%%%%%%%%%%%%%%%%%%%%%%%%%%%%%%%%%%
%%%%%%%%%%%%%%%%%%%%%%%%%%%%%%%%%%%%%%%%%%%%%%%%%%%%%%%%%%%%%%%%%%%%%%%%%%%%%%%

\bibliography{refs_NTK_DFL}
\bibliographystyle{icml2025}

\appendix
\onecolumn

\newpage
\appendix
\section{NTK-DFL Algorithms}
\label{appendix:algorithms}
\begin{algorithm}[ht]
   \caption{Consolidated Federated Learning Process}
   \label{alg:consolidated-fl-process}
   \begin{algorithmic}[1]
   \REQUIRE A set of clients $\mathcal{C}$ 
   \STATE Initialize weights $\vw_i^{(0)}$ for each client $i$.
   \FOR{each communication round $k = 1$ to $K$}
       \STATE Initialize graph structure $G^{(k)} = (\mathcal{C}, E^{(k)})$, specifying the neighbors $\mathcal{N}_i^{(k)}$ for each client $i$.
       \STATE Execute Algorithm \ref{alg:param-avg} for Per-Round Parameter Averaging
       \STATE Execute Algorithm \ref{alg:jacobian-computation-sending} for Local Jacobian Computation and Sending
       \STATE Execute Algorithm \ref{alg:weight-evolution} for Weight Evolution
   \ENDFOR
\end{algorithmic}
\end{algorithm}

\begin{algorithm}[ht]
   \caption{Per-Round Parameter Averaging}
   \label{alg:param-avg}
   \begin{algorithmic}[1]
   \REQUIRE For each client $i$, a set of neighbors $\mathcal{N}_i^{(k)}$ and initial weights $\vw_i^{(k)}$
      \FOR{each client $i \in \mathcal{C}$ \textbf{in parallel}}
         \STATE Send $\vw_i^{(k)}$ to all neighbors ${j \in \mathcal{N}_i^{(k)}}$
         \STATE Receive $\vw_j^{(k)}$ from all neighbors ${j \in \mathcal{N}_i^{(k)}}$
         \STATE $\bar{\vw}_i^{(k)} \gets  \frac{1}{{N_i + \sum_{j\in \mathcal{N}^{(k)}_i}N_j}} \big( {N_i\vw_i^{(k)} + \sum_{j \in \mathcal{N}_i^{(k)}} N_j\vw_j^{(k)} }\big)$
         \STATE Send aggregated weight $\bar{\vw}_i^{(k)}$ back to all neighbors ${j \in \mathcal{N}_i^{(k)}}$
      \ENDFOR
\end{algorithmic}
\end{algorithm}

\begin{algorithm}[ht]
  \caption{Local Jacobian Computation and Sending Jacobians}
  \label{alg:jacobian-computation-sending}
  \begin{algorithmic}[1]
  \REQUIRE Each client $i$ knows its neighbors $\mathcal{N}_i^{(k)}$ and has access to local data $\mathbf{X}_i$ and the aggregated weights $\bar{\vw}_j^{(k)}$ from each neighbor ${j \in \mathcal{N}_i^{(k)}}$.

     \FOR{each client $i \in \mathcal{C}$ \textbf{in parallel}}
        \STATE Compute the Jacobian $\mJ^{(k)}_{i,i} \equiv \nabla_{{\vw}} f(\mathbf{X}_i; \bar{\vw}_i^{(k)})$ using the client's own aggregated weight $\bar{\vw}_i^{(k)}$ and local data $\mathbf{X}_i$.
        \FOR{each neighbor ${j \in \mathcal{N}_i^{(k)}}$}
           \STATE Compute the Jacobian $\mJ^{(k)}_{i,j} \equiv \nabla_{{\vw}} f(\mathbf{X}_i; \bar{\vw}_j^{(k)})$ using the neighbor's aggregated weight $\bar{\vw}_j^{(k)}$ and client's local data $\mathbf{X}_i$.
           \STATE Send $\mJ^{(k)}_{i,j}$, true label $\mathbf{Y}_i$, and function evaluation $f(\mathbf{X}_i; \bar{\vw}_j^{(k)})$ to neighbor $j$.
        \ENDFOR
     \ENDFOR

\end{algorithmic}
\end{algorithm}
    
\begin{algorithm}[ht]
  \caption{Weight Evolution}
  \label{alg:weight-evolution}
  \begin{algorithmic}[1]
  \REQUIRE Each client $i$ has access to local data $\mathbf{X}_i$ and initial weights $\bar{\vw}_i^{(k)}$, and knows its neighbors $\mathcal{N}_i^{(k)}$
  \FOR{each client $i \in \mathcal{C}$ after intra-client communication}
      \STATE Compute local Jacobian tensor $\mJ_{i,i}^{(k)}$ and receive $\mJ_{j,i}^{(k)}$ from each neighbor $j$
      \STATE Construct tensor $\bm{\mathcal{J}}_i^{(k)}$ from $\{\mJ_{ii}^{(k)}\} \cup \{\mJ_{ji}^{(k)} \mid {j \in \mathcal{N}_i^{(k)}}\}$
      \STATE Compute local NTK $\mathbf{H}_i^{(k)}$ using $\bm{\mathcal{J}}_i^{(k)}$:
      \FOR{each data point pair $(x_m, x_n)$}
          \STATE $[\mathbf{H}^{(k)}_{i}]_{m,n} \gets \frac{1}{d_2} \langle \bm{\mathcal{J}}_i^{(k)}(x_m), \bm{\mathcal{J}}_i^{(k)}(x_n) \rangle_F$
      \ENDFOR
      \FOR{each timestep $t = 1$ to $T$}
          \STATE $\vf^{(k,t)}(\bm{\mathcal{X}}_i) \gets (\mathbf{I}-e^{-\frac{\eta t}{\tilde{N_i}}\mathbf{H}_i^{(k)}}) \bm{\mathcal{Y}}_i^{(k)} + e^{-\frac{\eta t}{\tilde{N_i}}\mathbf{H}_i^{(k)}} \vf^{(k)}(\bm{\mathcal{X}}_i)$
          \STATE $\bar{\vw}_i^{(k,t)} \gets \sum_{j=1}^{d_2} (\bm{\mathcal{J}}^{(k)}_{i,:j:})^T \mR_{i,:j}^{(k,t)} + \bar{\vw}_i^{(k)}$
      \ENDFOR
      \STATE Select $\vw_i^{(k+1,0)} \gets \bar{\vw}_i^{(k,t)}$ with the lowest loss given the residual $\vf^{(k,t)}(\bm{\mathcal{X}}_i) - \bm{\mathcal{Y}}_i$
  \ENDFOR
\end{algorithmic}
\end{algorithm}

\section{{Additional Details on Weight Evolution}}
% \CommentWong{Please color all newly added content (including the captions) and updated content (only the updated sentence, not the whole sentence).}
% \CommentGabe{Everything below this is new. I won't color it all for ease of readability.}
\label{appendix:C}
In implementation, computing the matrix exponential $e^{-\frac{\eta t}{\tilde{N_i}}\mathbf{H}_i^{(k)}}$ in  (\ref{alg:weight-evolution}) to evolve weights can be computationally expensive. In practice, the weights are evolved according to the more general differential equation from which (\ref{eq:analytical_dntk}) is derived, reliant upon the linearized model approximation $\vf({\bm{\mathcal{X}}_i};\bar{\vw}_j^{(k,t)}) \approx \vf(\bm{\mathcal{X}}_i;\bar{\vw}_j^{(k,0)})+\nabla_{{\vw}} \vf(\bm{\mathcal{X}}_i;\bar{\vw}_j^{(k,0)})^\top (\bar{\vw}_j^{(k,t)}-\bar{\vw}_j^{(k,0)})$. The differential equation is as follows
\begin{equation}
\frac{d}{dt}\vf(\bm{\mathcal{X}}_i;\bar{\vw}_j^{(k,t)}) = -{\eta}\mathbf{H}_j^{(k)}\nabla_{\vf}\mathcal{L}.   
\end{equation}
Here, $\mathcal{L}$ is the loss function. For example, for a half mean-squared error (MSE) loss, term on the right becomes the residual matrix $\nabla_{\vf}\mathcal{L} = \vf(\bm{\mathcal{X}}_i;\bar{\vw}_j^{(k,t)}) - \bm{\mathcal{Y}}_i$. During weight evolution, a client $j$ evolves their neighboring function evaluation from the initial condition $\vf(\bm{\mathcal{X}}_i;\bar{\vw}_j^{(k,0)})$ to the time-evolved $\vf(\bm{\mathcal{X}}_i;\bar{\vw}_j^{(k,t)})$ using a differential equation solver and the differential equation above. To implement (\ref{eq:weights_dntk}), we use a process similar to \citet{yue2022neuraltangentkernelempowered} where the initial client residual is evolved over a series of timesteps specified by the user. For user-specified timesteps, the loss at that time is found using the evolved residual. Then, the best-performing weights are evolved using the left side of (\ref{eq:weights_dntk}) and selected for the next communication round.

\section{Additional Experimental Details}
\label{appendix:B}
\subsection{Baselines}
\label{appendix:baselines}
NTK-FL is the only centralized baseline that we compare with. We choose a per-round client fraction that ensures that the busiest node in the centralized setting is no busier than the busiest decentralized setting. By busier, we mean the degree of the node or the number of clients communicating with it. We note that NTK-FL is not an upper bound in this case due to the comparison being founded on node busyness, which disadvantages a centralized approach where all communication happens through a single, centralized node. Evaluating NTK-FL in the same setting as  the table in Figure \ref{fig:fashion_mnist_convergence}, NTK-FL converges to threshold accuracy in 73, 85, and 180 communication rounds for heterogeneity settings IID, $\alpha=0.5$, and $\alpha=0.1$, respectively. D-PSGD \citep{d-psgd} is one of the first decentralized, parallel algorithms for distributed machine learning that allows nodes to only communicate with neighbors. DFedAvg \citep{dfedavg} adapts FedAvg to the decentralized setting, and DFedAvgM makes the use of SGD-based momentum and extends DFedAvg. Both use multiple local epochs between communication rounds, like vanilla FedAvg. DFedSAM \citep{dfedsam} incorporates the SAM algorithm \citep{sam} into the DFL process. DisPFL \citep{dispfl} is a personalized federated learning approach that aims to train a global model and personalize it to each client with a local mask. In order to make the comparision fair, we report the accuracy of the global model on our test set.

\subsection{Hyperparameters}
\label{appendix:baseline-hyperparameters}
We perform a hyperparameter search over each baseline and select the hyperparameters corresponding to the best validation accuracy. We use the $\alpha=0.1$ Fashion-MNIST test accuracy at communication round 30 as the metric for selection. This is done because the majority of comparisons take place on Fashion-MNIST in the non-IID setting. For D-PSGD, we use a learning rate of $0.1$, and a batch size of $10$ (local epoch count is defined to be one in this approach). For DFedAvg, we use a learning rate of $0.1$, a batch size of $25$, and $20$ local epochs. For DFedAvgM, we use a learning rate of $0.01$,  a batch size of $50$, $20$ local epochs, and a momentum of $0.9$. For DisPFL, we use a learning rate of $0.1$, a batch size of $10$, and $10$ local epochs. Following \citet{dispfl}, we use the sparsity rate of $0.5$ for DisPFL. For DFedSAM, we began with the parameters suggested in \citet{dfedsam}. After a hyperparameter search, we found that a radius $\rho=0.01$, $\eta=0.01$, momentum of $0.99$, learning rate decay of $0.95$, weight decay of $5\times10^{-4}$, 5 local epochs, and a batch size of 32 yielded the best performance. Note that we used a single gossip step per round for all approaches in order to maintain a fair comparison. As for the NTK-DFL, we use a learning rate of 0.01 and search over values $t\in\{100, 200, \ldots, 800\}$ during the weight evolution process. 
For the baseline compression methods, we experimented with quantization as well as top-$k$ sparsification~(see Appendix \ref{appendix:mitigation}). We found that the baseline methods, which were gradient-based, cannot withstand the aggressive sparsification applied to NTK-DFL. Therefore, we used only quantization [such as in \citet{dfedavg}] and selected the most effective value to facilitate a fair comparison.

\subsection{Inter-Client Model Deviation}
\label{subsec:inter-client-def}
We plot the relationship between the accuracy and the inter-client model deviation among NTK-DFL clients in Figure \ref{fig:var_vs_acc}, where the inter-client model deviation is defined for a set of weight vectors $\vw_1, \ldots, \vw_M\in\mathbb{R}^{d}$ as follows:
\begin{equation}
    V = \frac{1}{d}\sum_{j=1}^{d} \sqrt{\sum_{i=1}^M (w_{i,j}-\bar{w}_j)^2},\quad \bar{\vw} = \frac{1}{M}\sum_{i=1}^M \vw_i, 
    \label{eq:inter-client-deviation}
\end{equation}
where it captures per-parameter variation in an averaged sense.

\subsection{Additional Experimental Results}
\label{appendix:extra-results}
\begin{figure*}[h]
    \centering
    \begin{minipage}[b]{0.49\textwidth}
        \centering
        {\includegraphics[width=0.98\linewidth]{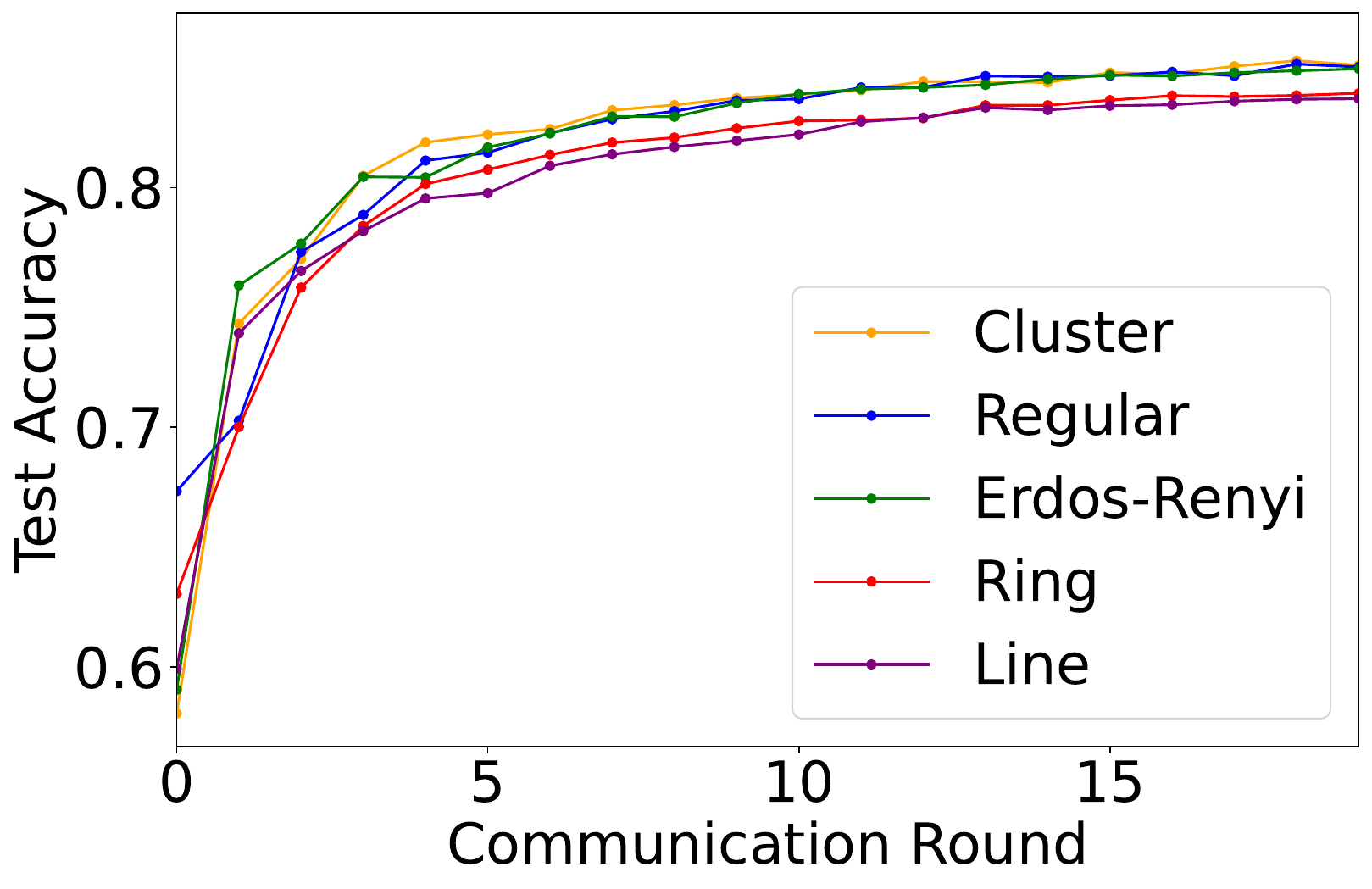}\vspace{-4mm}}
        \caption{Convergence of NTK-DFL across different dynamic topologies, trained on Fashion-MNIST. NTK-DFL is evaluated with {a clustered graph (in yellow) with 5 neighbors per client}, a $\kappa=5$ regular graph (in blue), an Erdos-Renyi random graph with five mean neighbors (in green), a ring topology (in red), and {a line topology (in purple)}. We observe that NTK-DFL demonstrates steady convergence across different topology classes. {For the ring and line topologies, convergence is a bit slower due to a sparser graph of 2 rather than 5  neighbors per client}. }
        \label{fig:all topology}
    \end{minipage}
    \hfill
    \begin{minipage}[b]{0.49\textwidth}
        \centering
        {\includegraphics[width=1.0\linewidth]{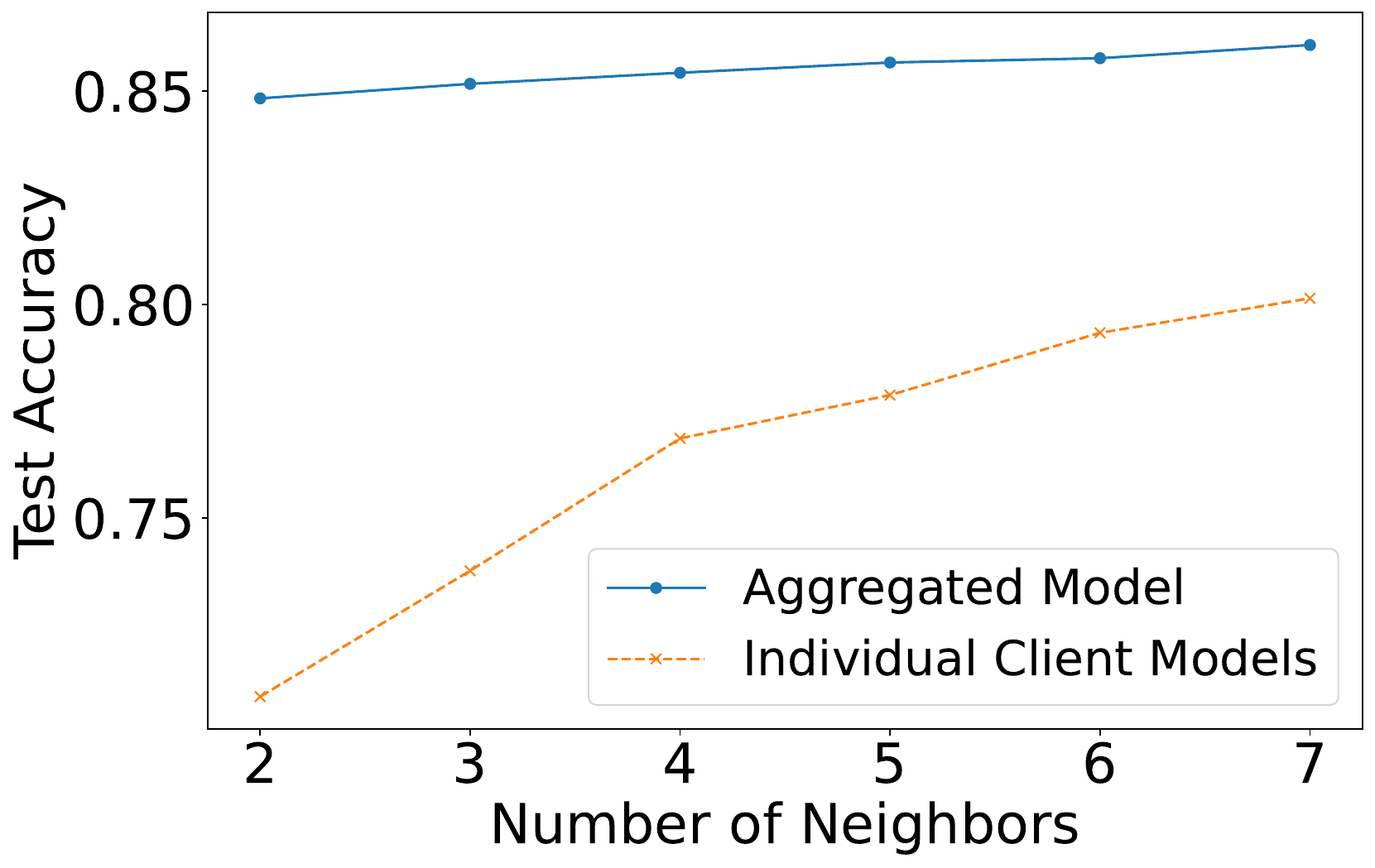}\vspace{-4mm}}
        \caption{NTK-DFL model accuracy as a function of neighbor count~$\kappa$, trained on Fashion-MNIST. Notably, the aggregated model accuracy across NTK-DFL clients (in blue) remains consistent, even as network sparsity varies. This stability persists despite a significant decline in mean individual client test accuracy (in yellow) as the number of neighbors decreases.\vspace{11mm}}
        \label{fig:acc_vs_sparsity_level}
    \end{minipage}
\end{figure*}

\begin{figure*}[h]
\centering
  {\begin{subfigure}[b]{0.3605\textwidth}
    \centering
    \includegraphics[width=\textwidth]{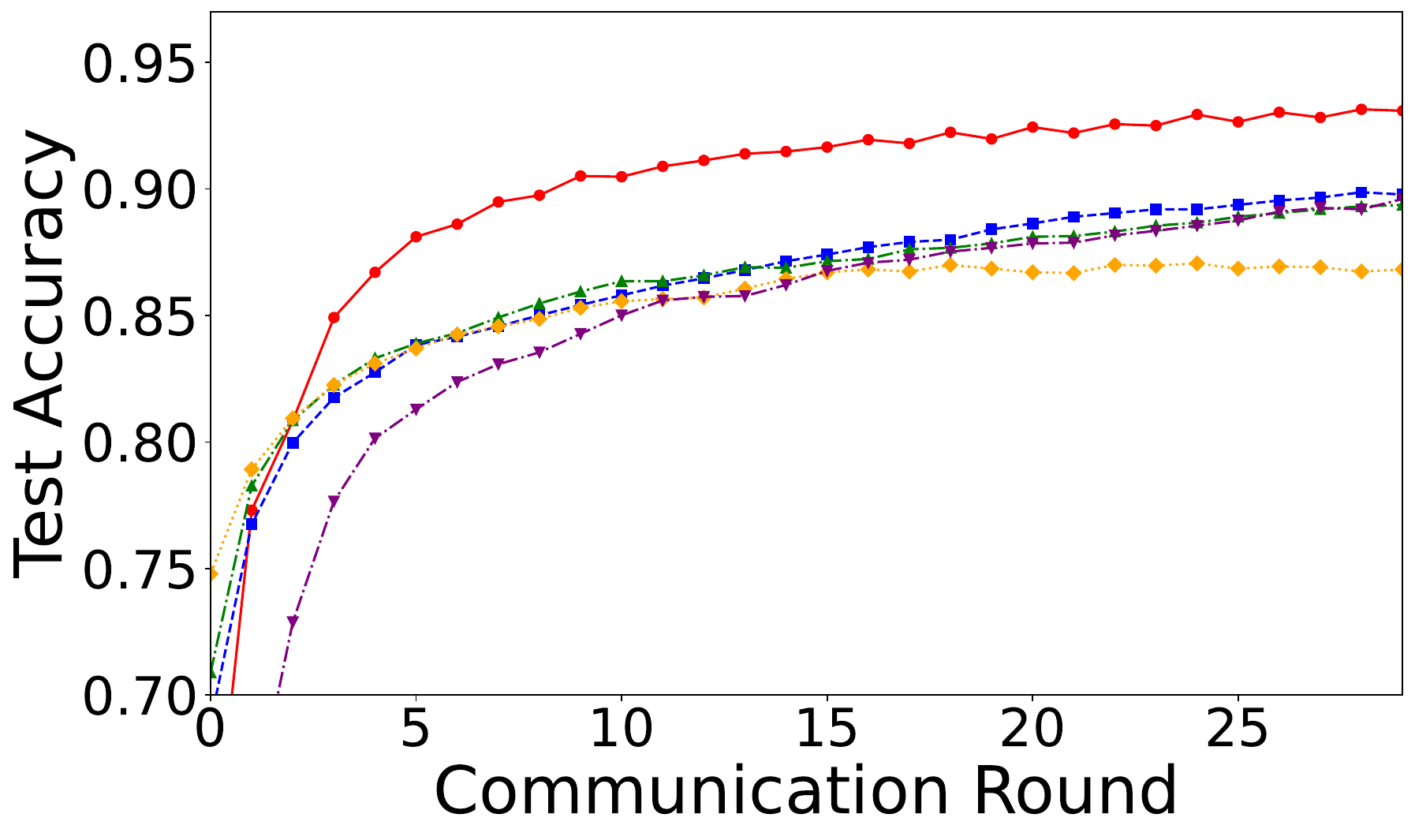}
    \subcaption{FEMNIST (Feature-skewed)}
  \end{subfigure}
  \begin{subfigure}[b]{0.314\textwidth}
    \centering
    \includegraphics[width=\textwidth]{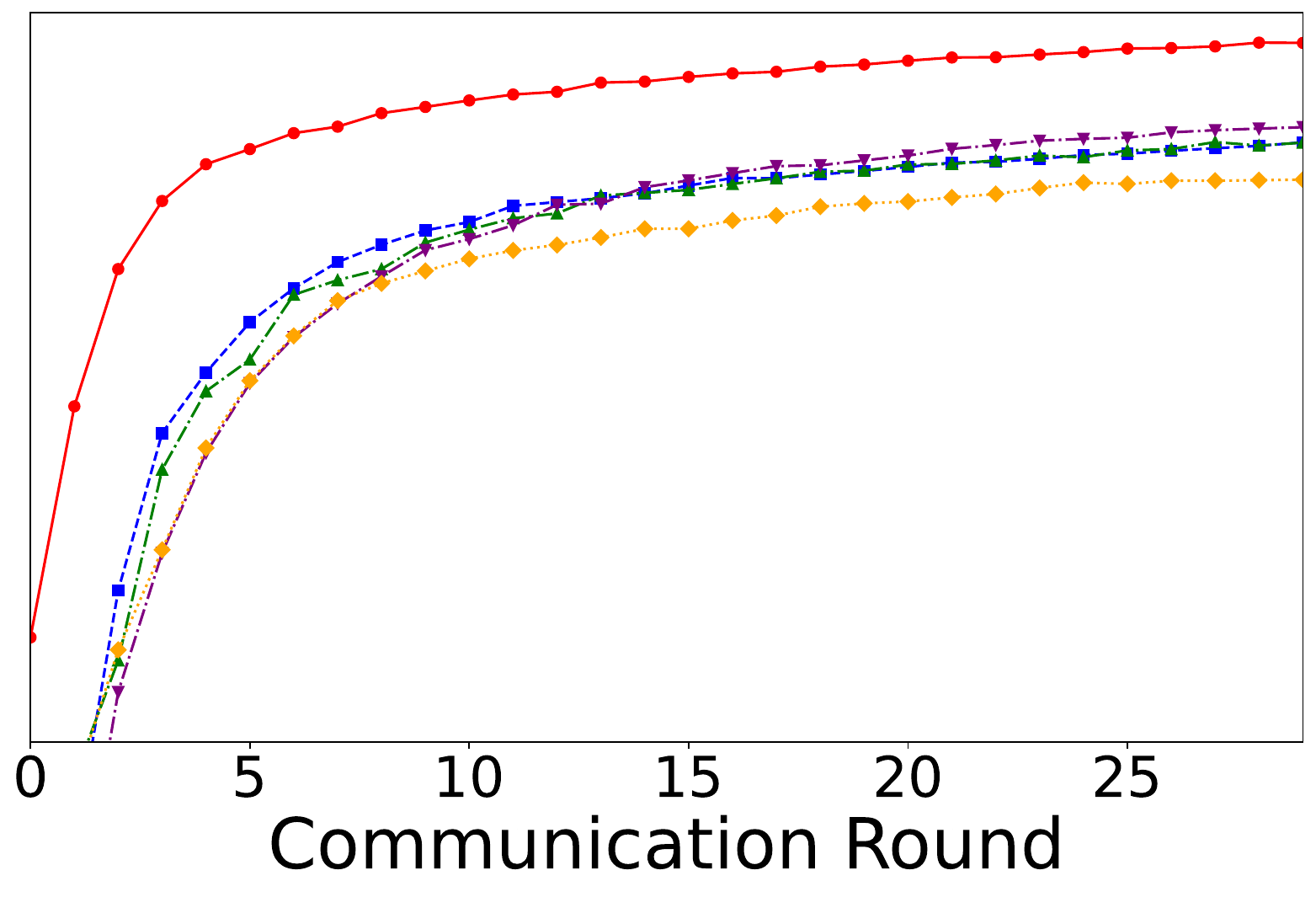}
    \subcaption{MNIST $(\alpha = 0.05)$}
  \end{subfigure}
    \begin{subfigure}[b]{0.3140\textwidth}
    \centering
    \includegraphics[width=\textwidth]{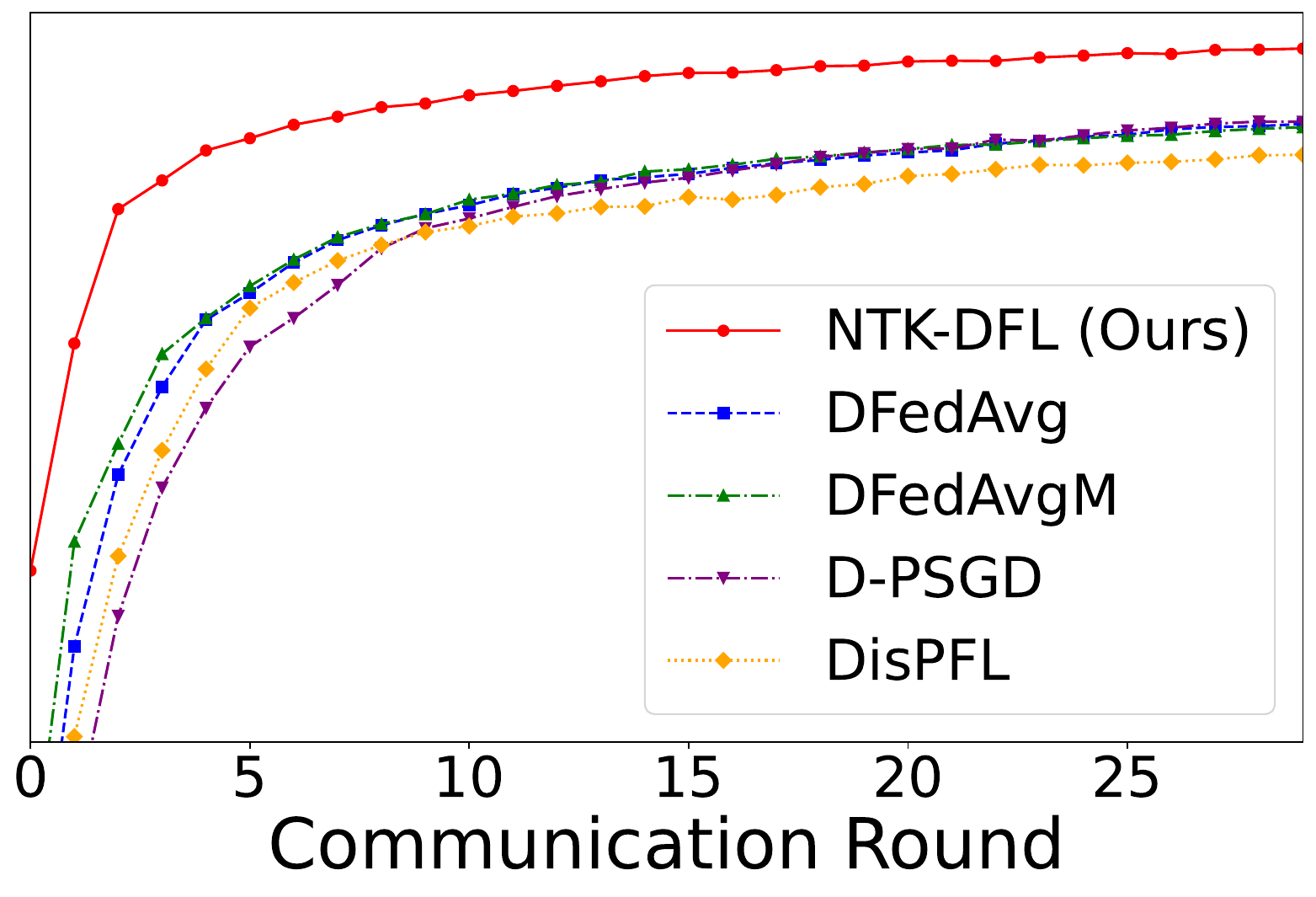}
    \subcaption{MNIST $(\alpha = 0.1)$}
  \end{subfigure}\vspace{-4mm}}
\caption{
Convergence of various methods on heterogeneous datasets: (a) FEMNIST, (b) Non-IID MNIST ($\alpha=0.05$), and (c)~Non-IID MNIST ($\alpha = 0.1$). NTK-DFL consistently outperforms all baselines. 
}
\label{fig:different_dataset_convergence}
\end{figure*}

\begin{figure*}[h]
    \centering
    \begin{minipage}[b]{0.49\textwidth}
        \centering
        {\includegraphics[width=\linewidth]{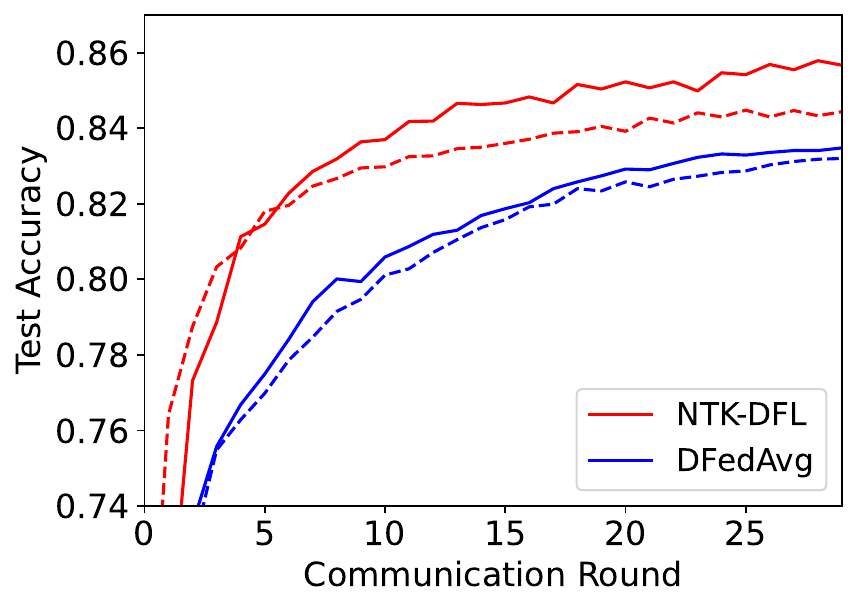}\vspace{-4mm}}
        \caption{The effect of static vs. dynamic topology on NTK-DFL (Fashion-MNIST, $\alpha=0.1$). 
        Solid lines correspond to a dynamic topology, whereas dotted lines correspond to a static topology. 
        Both methods benefit from the dynamic topology and NTK-DFL outperforms DFedAvg under both topologies.
        Other baselines are not drawn but perform similarly to DFedAvg.}
        \label{fig:dynamic}
    \end{minipage}
    \hfill
    \begin{minipage}[b]{0.49\textwidth}
      \centering
      {\includegraphics[width=\linewidth]{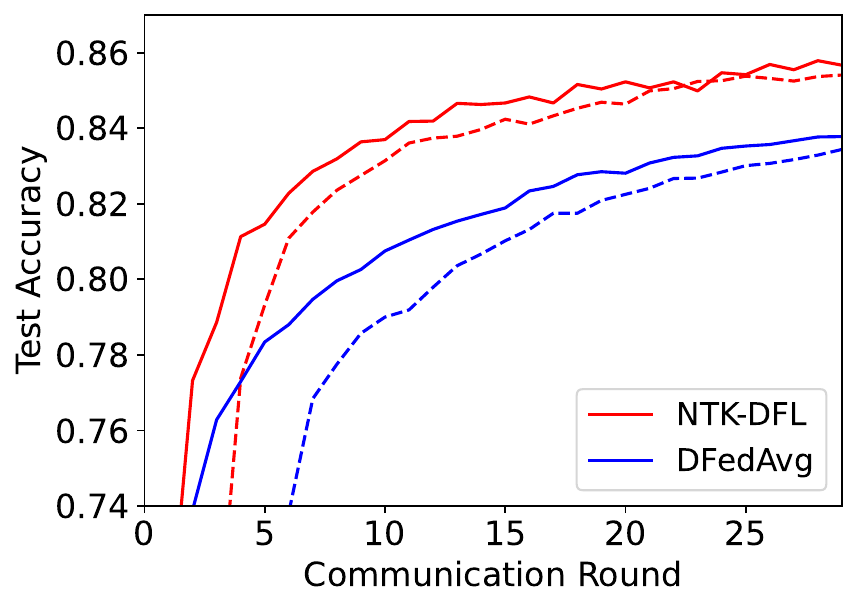}\vspace{-4mm}}
      \caption{
      The effect of different vs. identical weight initialization (Fashion-MNIST, $\alpha=0.1$).
      Solid lines correspond to the same weight initialization for all clients, whereas dotted lines correspond to different initialization.
      The convergence of NTK-DFL is affected less than that of DFedAvg.
      Other baselines are not drawn but perform similarly to DFedAvg.}
      \label{fig:weight init}
    \end{minipage}
\end{figure*}

\section{Discussion of Communication Trade-offs}
\label{appendix:tradeoffs}
Federated learning methods balance two key communication dimensions: the volume of data transmitted per round and the total number of communication rounds required. NTK-DFL transmits more data per round than traditional approaches such as DFedAvg by sharing Jacobian matrices rather than gradients or weights, but requires fewer total rounds. 
We emphasize that the number of communication rounds is a key practical consideration. Below, we highlight scenarios where minimizing it is especially beneficial:

\begin{itemize}
    \item \textbf{Large per-round communication latency}, where fewer rounds can significantly reduce overall training time (e.g., compression or encryption of weights/gradients, heavy preprocessing of input data).
    \item \textbf{Limited device availability}, in which fewer rounds allow more efficient training when devices are intermittently available.
    \item \textbf{High bandwidth applications}, where ample network bandwidth (e.g., gigabit home internet) can accommodate a large data volume for each communication round, making the number of communication rounds the dominant factor in training efficiency.
    \item \textbf{Synchronization delays}, where each round must wait for all devices to complete computation, with the slowest device bottlenecking progress, thus making the number of communication rounds an important factor.
\end{itemize}

\section{{Overhead Mitigation Strategies}}
\label{appendix:mitigation}

While analysis of memory and communication overhead are not a central theme of this paper, we include strategies to mitigate both forms of overhead for practical deployment. A thorough analysis of optimization and parallelization is out of the scope of this work and we leave it to future research.

\subsection{{Jacobian Batching}}
\begin{figure}
    \centering
    \begin{minipage}[t]{0.49\textwidth}
        \centering
        {\includegraphics[width=\linewidth]{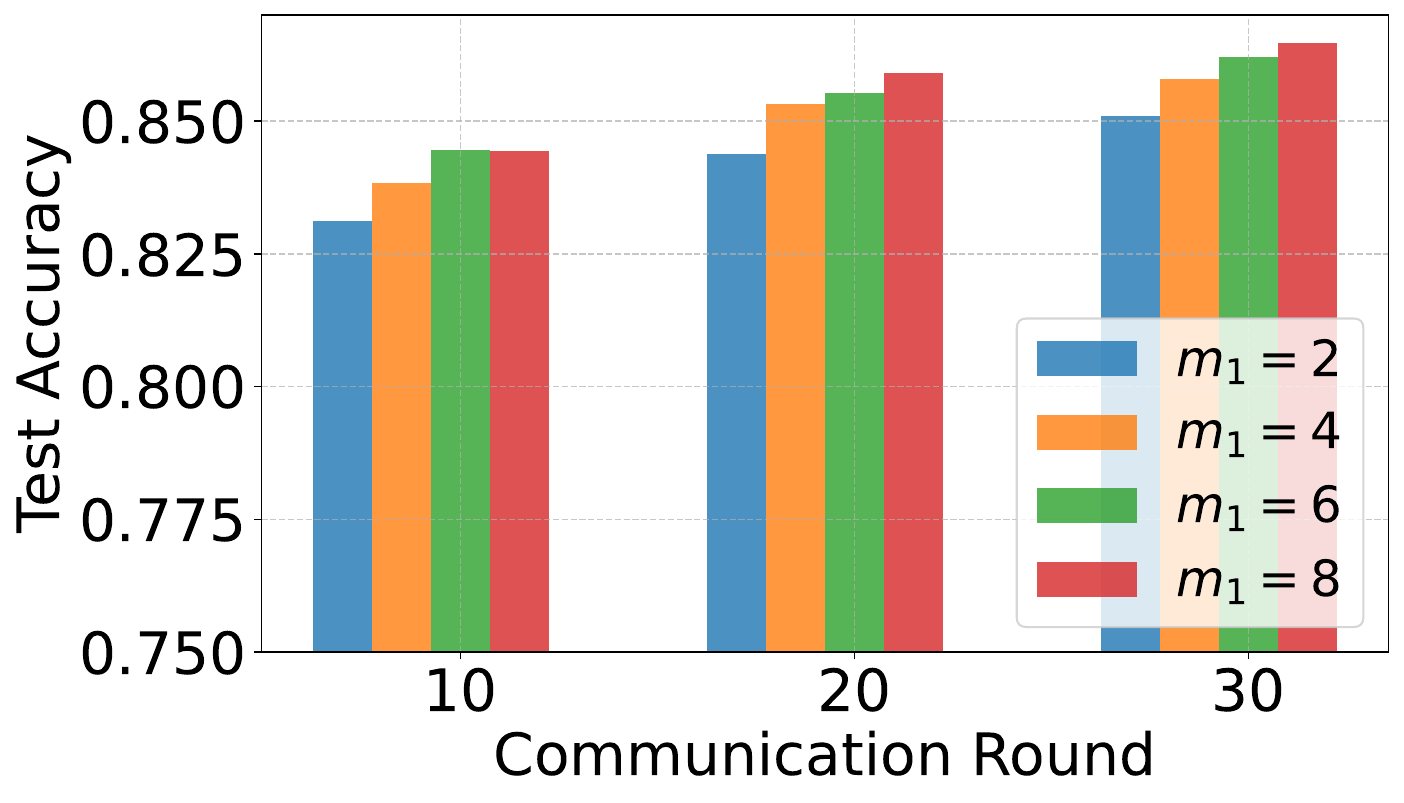}\vspace{-4mm}}
        \caption{Test accuracy of NTK-DFL vs. communication round for various Jacobian batch numbers $m_1$, with higher $m_1$ values denoting more batches (Fashion-MNIST, $\alpha=0.1$). We observe a general, counterintuitive increase in test accuracy with an increased number of batches.}
        \label{fig:jacobian_batching_compare}
    \end{minipage}
    \hfill
    \begin{minipage}[t]{0.49\textwidth}
      \centering
      {\includegraphics[width=\linewidth]{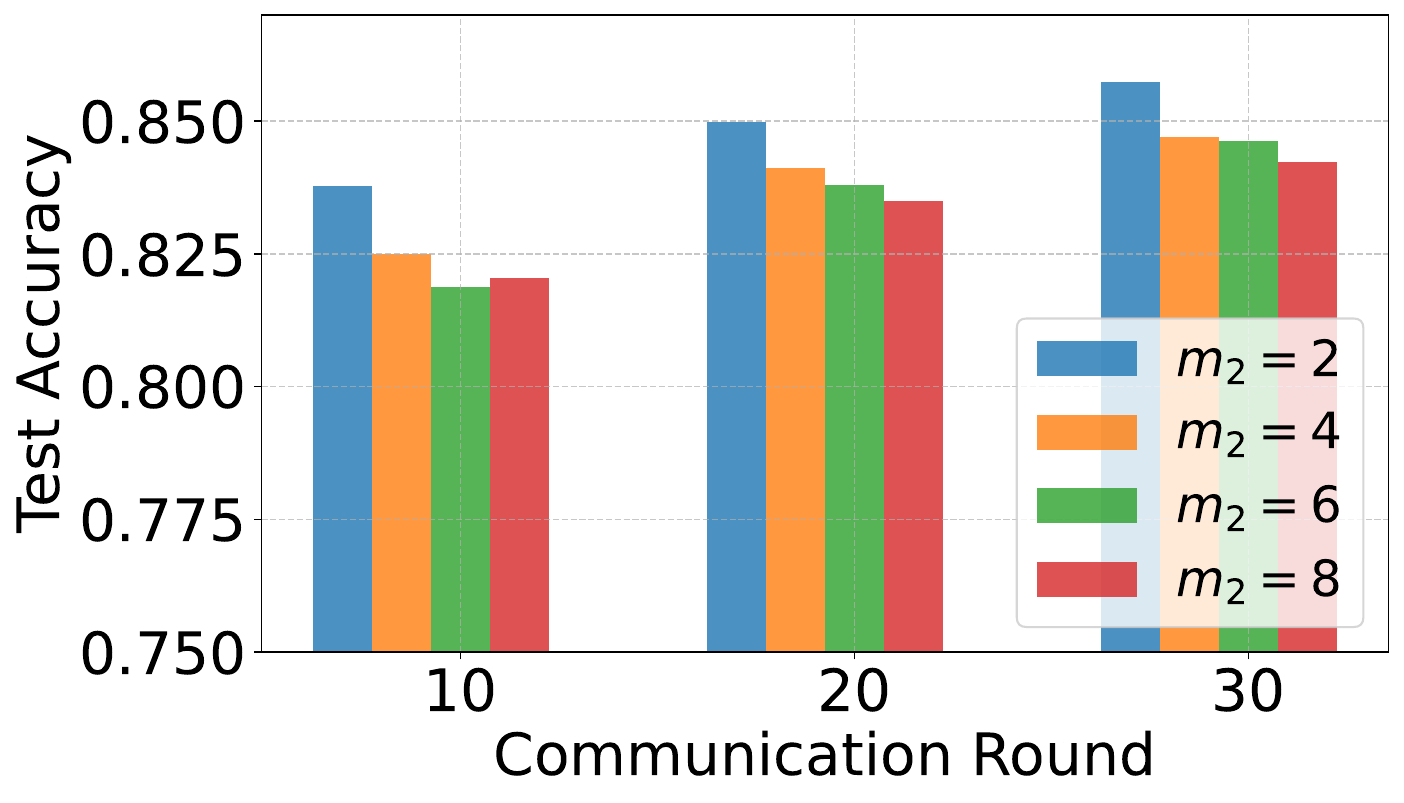}\vspace{-4mm}}
      \caption{Test accuracy of NTK-DFL vs. communication round for sampling divisors $m_2$ (Fashion-MNIST, $\alpha=0.1$). Different from Jacobian batching, only a $1/m_2$ fraction of client data is selected each communication round. We observe a slight decrease in test accuracy with increased $m_2$.}
      \label{fig:sampling}
    \end{minipage}
\end{figure}

We introduce Jacobian batching to address potential memory constraints in NTK-DFL implementations. For scenarios involving dense networks or large datasets, clients can process their local datasets in smaller batches, reducing memory complexity from $O(N_id_2d)$ to $O(N_id_2d/m_1)$, where $m_1$ is the number of batches. Clients compute and transmit Jacobians for each batch separately, evolving their weights multiple times per communication round. This approach effectively trades a single large NTK $\mathbf{H}\in\mathbb{R}^{N\times N}$ for $m_1$ smaller NTKs $\mathbf{H}_{m_1}\in\mathbb{R}^{N/{m_1}\times N/{m_1}}$ that form block diagonals of $\mathbf{H}$, where $N$ represents the total number of data points between client $i$ and its neighbors $\mathcal{N}_i$. While some information is lost in the uncomputed off-diagonal entries of $\mathbf{H}$, this is mitigated by the increased frequency of NTK evolution steps. Figure~\ref{fig:jacobian_batching_compare} demonstrates this phenomenon, where an increasing batch number $m_1$ actually leads to improved convergence. This complexity reduction enables clients to connect in a denser network for the same memory cost. 

\subsection{Use of the Clustered Topology}
\label{section:clustered}
In the traditional NTK-DFL approach, each client must compute Jacobians on their own data with respect to the weight vector of every neighbor, as NTK construction requires all Jacobians to be evaluated at the same point in weight space. This implies that Jacobian computation scales linearly with the number of neighbors. However, a (dynamic) \textit{clustered topology} reduces this burden: after the initial weight synchronization step, all clients within a cluster share the same (aggregated) weight, so each client only needs to compute a single set of Jacobians. These Jacobians can then be reused for all neighbor interactions within the cluster. Furthermore, the weight evolution step can be offloaded to a single designated client, or distributed across clients by sharing intermediate weights during the weight unrolling process~(\ref{eq:weights_dntk2}). As a result, \textbf{both communication and computation overheads become independent of the number of neighbors}, assuming each round's evolution is handled by a single client per cluster. We illustrate the performance of NTK-DFL under this clustered topology in Figure \ref{fig:all topology} of Appendix \ref{appendix:B}, which shows convergence comparable to that of a regular graph with the same average degree.

\subsection{{Communication Cost}}
Compared to traditional weight-based approaches that communicate a client's parameters $\vw_i$ each round, NTK-DFL utilizes Jacobian matrices to enhance convergence speed and heterogeneity resilience. This tensor has memory complexity $O(N_id_2d)$, where $N_i$ denotes the number of data points between client $i$ and its neighbors $\mathcal{N}_i$, $d$ is the model parameter dimension, and $d_2$ is the output dimension. We propose the following strategies to improve the communication efficiency of NTK-DFL while maintaining convergence properties in heterogeneous settings. 

\textbf{Data Subsampling}\quad We introduce an approach where clients sample a $1/{m_2}$ 
% \CommentWong{Please use another symbol to avoid symbol collision. You used $m$ in D.1. You can use $m_1$ and $m_2$ for D.1 and D.2, respectively.} 
fraction of their data each round for NTK evolution. Clients follow the protocol described in Section \ref{section:Communication Protocol}, but exchange Jacobian matrices of reduced size. As demonstrated in Figure~\ref{fig:sampling}, moderate values of $m$ yield light performance degradation, validating this communication reduction strategy.

\begin{figure}
    \centering
    \begin{minipage}[b]{0.50\textwidth}
        \centering
        {\includegraphics[width=\linewidth]{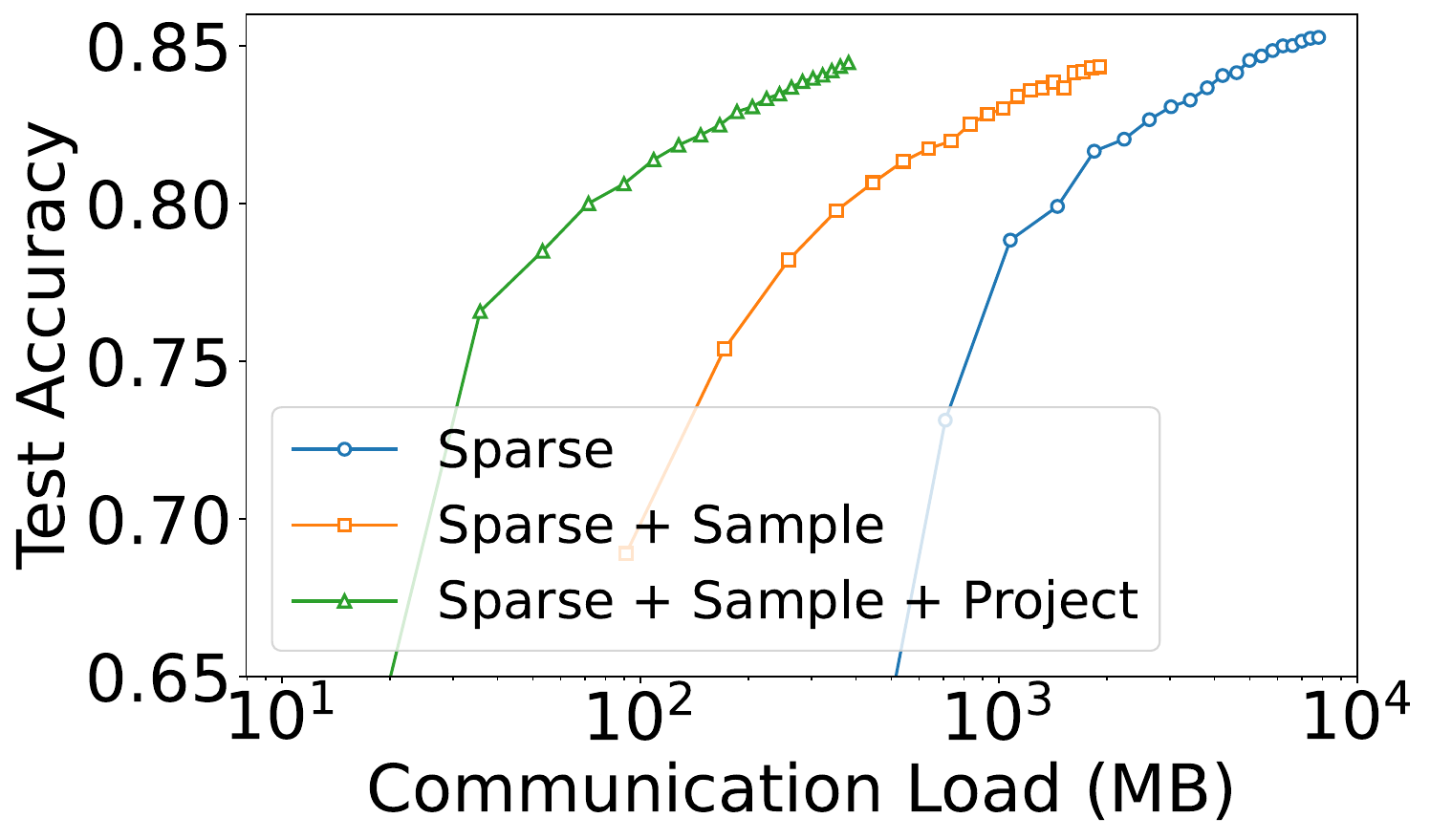}\vspace{-4mm}}
        \caption{Comparison of NTK-DFL variants with progressive communication optimizations (Fashion-MNIST, $\alpha=0.1$). Data sampling and projection technique provides compounding reductions in communication load compared to sparsification alone, while the fully optimized variant demonstrates significantly lower communication requirements at a comparable test accuracy.
        % \CommentWong{Please use dots or circles on the plot to show the exact operating points you experimentally obtained.}
        }
        \label{fig:communication_comparison}
    \end{minipage}
    \hfill
    \begin{minipage}[b]{0.44\textwidth}
      \centering
      {\includegraphics[width=\linewidth]{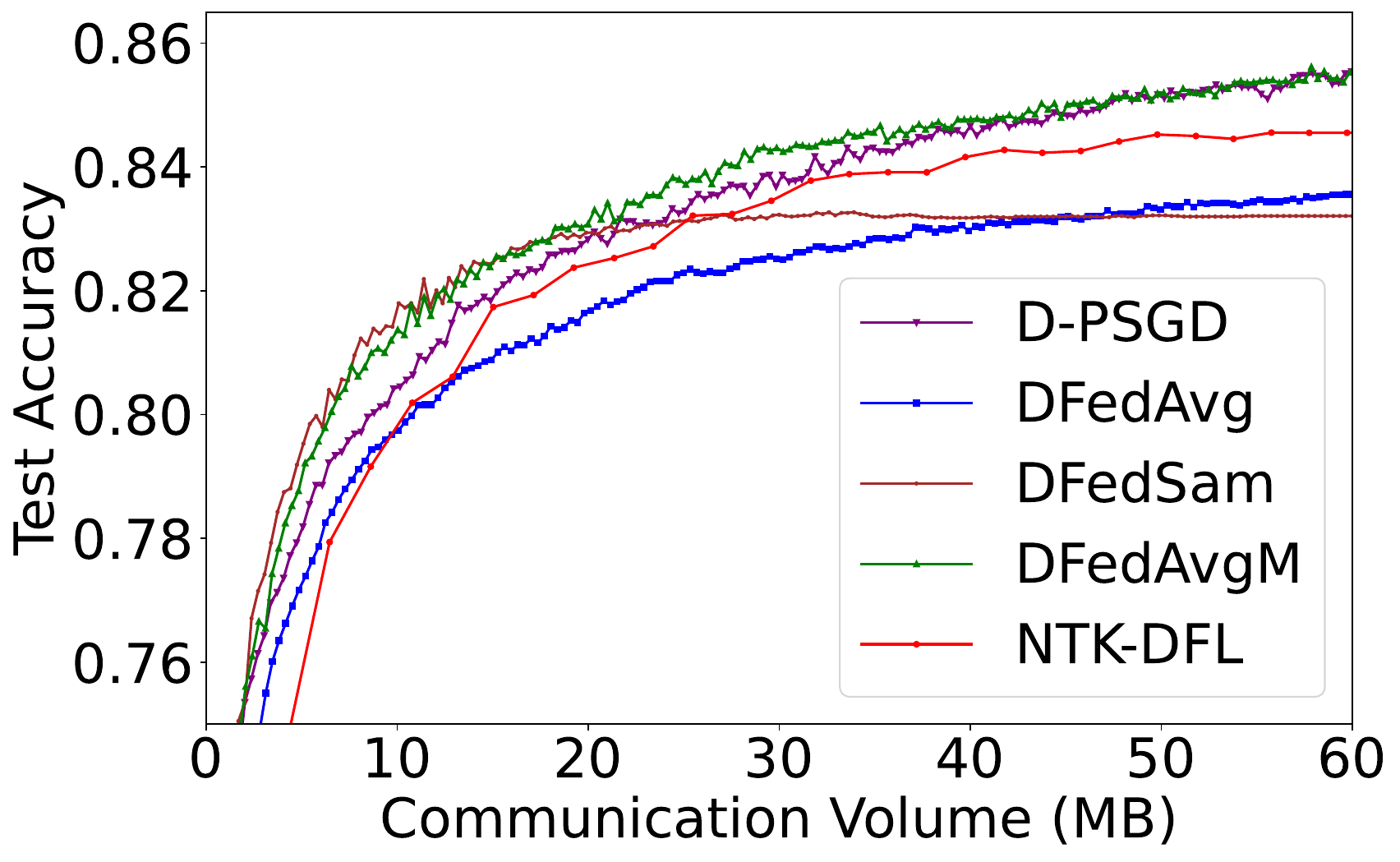}\vspace{-4mm}}
      \caption{
      Test accuracy vs. communication volume for NTK-DFL and baselines (Fashion-MNIST, $\alpha=0.1$). 
      % While NTK-DFL achieves convergence in fewer communication rounds than DFedAvg, its more expressive parameter updates require a higher communication volume per round.
      With the use of compression techniques and a clustered topology, NTK-DFL performs similarly in term of communication volume. It is slightly outperformed by more communication efficient algorithms, though they need about 5 times as many communication rounds to converge.
      }
      \label{fig:comm_compare_bits}
    \end{minipage}
\end{figure}

\textbf{Jacobian Compression}\quad We employ several techniques to reduce Jacobian tensor dimensionality. First, we apply top-$k$ sparsification, zeroing out elements with the smallest magnitude \citep{topk}. The remaining nonzero values are quantized to $b$ bits. Additionally, we introduce a shared random projection matrix $\mathbf{P}\in\mathbb{R}^{d_1 \times d_1'}$ generated from a common seed, creating projections $\mathbf{Z}_i = \mathbf{X}_i\mathbf{P}$ that reduce input dimension from $d_1$ to $d_1'$. This combination of techniques maintains convergence properties while significantly reducing communication costs. Note that similar compression schemes applied to weight-based approaches lead to significant degradation in performance \citep{yue2022neuraltangentkernelempowered}. Figure \ref{fig:communication_comparison} illustrates the relative differences in communication load for a different combinations of the techniques above, with a sparsification of 0.5, quantization to 6 bits, a sampling of $m_2=5$, and a projection to $d_1' = 200$ for the full optimization curve. 

% Figure \ref{fig:comm_compare_bits} compares NTK-DFL updates with less expressive weight updates of the baseline methods. NTK-DFL uses the compression methods detailed above, as well as a clustered topology (Section \ref{section:clustered}) to reduce communication volume. The communication-optimized NTK-DFL converges in fewer communication rounds than the baselines. However, with more expressive updates than DFedAvg, it uses greater communication volume. This enforces the idea that NTK-DFL is especially useful in scenarios where convergence in few communication rounds is important, such as those with non-negligible encoding and decoding delays. 

Figure \ref{fig:comm_compare_bits} compares the convergence of NTK-DFL with the baseline methods in terms of communication volume. NTK-DFL uses the compression methods detailed above, as well as a clustered topology (Section \ref{section:clustered}) to reduce communication overhead. The communication-optimized NTK-DFL converges in fewer rounds than the baselines. However, with more expressive updates, it uses greater communication volume. This enforces the idea that NTK-DFL is especially useful in scenarios where convergence in fewer 
% \CommentWong{a few or fewer?}
rounds is important, such as those outlined in Appendix~\ref{appendix:mitigation}.

% \EditGabe{We note that we apply quantization to the baseline methods to facilitate a fair comparison, performing a hyperparameter sweep over quantization levels to select the most effective level. The baseline methods, which are gradient-based and less expressive, do not converge under the aggressive sparsification described above.}

\begin{figure*}[!b]
    \centering
    {\includegraphics[width=0.8\linewidth]{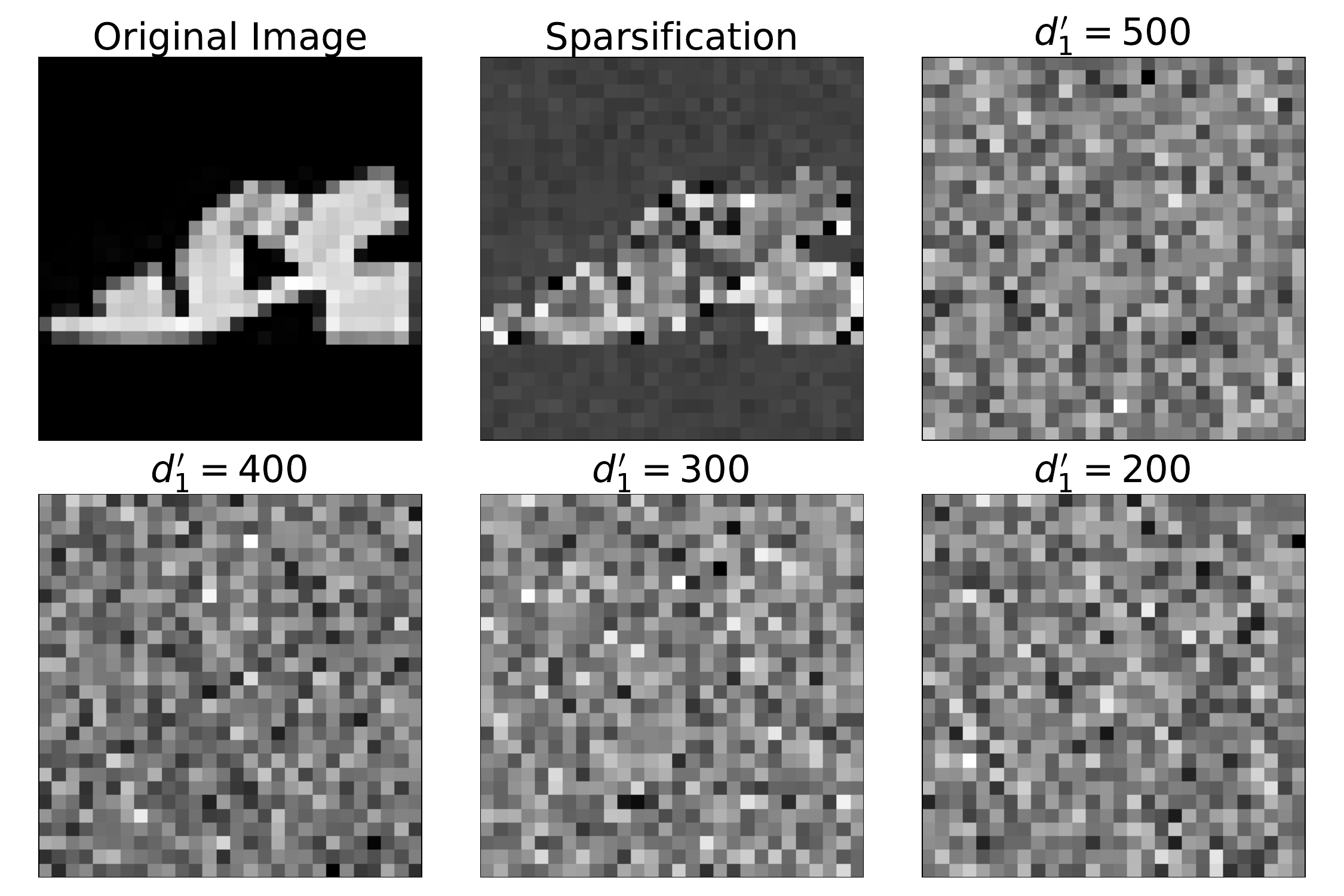}
    \vspace{-4mm}}
    \caption{Reconstruction attack of client data from Jacobian matrices for various levels of compression. For the image corresponding to sparsified matrices (the middle image of the first row), no random projection is done. We observe the ability to reconstruct a very noisy version of client data. For the other images, we use sparsification and a random projection to dimension $d_1'$. We observe an inability to reconstruct client data when the random projection is additionally applied.\vspace{-0mm}}
    \label{fig:privacy_reconstruction}
\end{figure*}

\section{Reconstruction Attack}
\label{appendix:recon-attack}
While privacy preservation is not the primary focus of this work, we conduct a brief analysis of data privacy in NTK-DFL. Following the reconstruction attack method of \citet{deep_leakage}, we evaluate the feasibility of reconstructing client data from transmitted Jacobian matrices under varying compression levels. Our experiments range from basic top-$k$ sparsification with sparsity $0.25$ to combined sparsification with random projection to dimension $d_1'=200$.  Figure~\ref{fig:privacy_reconstruction} illustrates that client data reconstruction becomes increasingly difficult when a random projection is additionally applied to the Jacobian matrices.

\clearpage
\section{Mathematical Analysis}
\label{appendix:math}
Below, we include more detailed assumptions that are used throughout the proof of the lemmas presented below and the subsequent proof of Theorem \ref{thm:main}. We note that the assumptions listed below are relatively mild and common in DFL settings \citep{dfedavg, dfedsam}.
\begin{assumption}
\label{assum:smooth}
(Lipschitz smoothness). The function \( \mathcal{L}_i \) is differentiable, and its gradient \( \nabla \mathcal{L}_i \) is \( L \)-Lipschitz continuous for all \( i \in \{1,2,\dots,m\} \), i.e.,
\begin{equation}
\|\nabla \mathcal{L}_i(\vw) - \nabla \mathcal{L}_i({\vv})\| \leq L \|\vw - \vv\|,
\end{equation}
for all \( \vw, \vv \in \mathbb{R}^d \).
\end{assumption}
\begin{assumption}
\label{assum:bounded variance}
(Bounded variance). The global variance of the gradient for functions \( \{\mathcal{L}_i\}_{i=1}^{m} \) is jointly bounded, i.e.,
\begin{equation}
\frac{1}{m} \sum_{i=1}^{m} \|\nabla \mathcal{L}_i(\vw) - \nabla \mathcal{L}(\vw)\|^2 \leq \sigma_g^2,
\end{equation}
for all \( \vw \in \mathbb{R}^d \). 
% We note that there is no local variance  $\sigma_l^2$ introduced by stochastic update steps in the vanilla NTK-DFL. 
This quantity is directly related to data heterogeneity among clients.
\end{assumption}
\begin{assumption}
\label{assum:bounded-gradient}
(Bounded gradients). For any \( i \in \{1,2,\dots,m\} \) and \( \vw \in \mathbb{R}^d \), the gradient of the local loss function is bounded as 
\(
\|\nabla \mathcal{L}_i(\vw)\| \leq B, 
\)
for some constant \( B > 0 \).
\end{assumption}
We present the following lemmas to aid in the analysis. {First, we present Lemma \ref{lem:mixing-projection} that is common in DFL literature relating to the spectral properties of the mixing matrix. The following lemmas bound the mean divergence of the client weights over local update steps $t$ (Lemma \ref{lem:per-step-difference}), the variance of client weights (Lemma \ref{lem:dev_from_avg}), and the mean gradient norm over clients (Lemmma \ref{lem:avg-clients-grad}). Lastly, we bound the error term $\delta_{\text{NTK}}$ (Lemma \ref{lem:delta_ntk_bound}).}
\input{tex_math/lemma1}

\input{tex_math/lemma2}
\input{tex_math/lemma2_proof}
\\
\input{tex_math/lemma3}

\input{tex_math/lemma4}

\input{tex_math/lemma5}
\input{tex_math/lemma5_proof}
\input{tex_math/lemma6}

\textbf{Proof of Theorem \ref{thm:main}.}

We begin with 
    \begin{equation}
    \bar{\vw}^{(k+1)}-\bar{\vw}^{(k)} = \bar{\vw}^{(k,T)}-\bar{\vw}^{(k)},
    \end{equation}
due to the fact that the application of the mixing matrix does not change $\bar{w}$. Using the NTK-DFL weight update step, noting that the average weight at round $k$ and time step $t=0$ is written as $\bar{\vw}^{(k)}\coloneqq\bar{\vw}^{(k,0)}$,
\begin{equation}
\bar{\vw}^{(k,T)}-\bar{\vw}^{(k)} = -\frac{\eta}{m}\sum_{i=1}^m\sum_{u=1}^T \nabla \mathcal{L}^{\text{NTK}}(\vw_i^{(k,u)}) =-\frac{\eta}{m}\sum_{i=1}^m\sum_{u=1}^T  \Bigl( 
\nabla_{\vw} \mathcal{L}_i (\vw_i^{(k,u)}) +
\Delta_{i,\text{NTK}}^{(k,u)}
\Bigr).
\end{equation}
Assumption \ref{assum:smooth} gives us
\begin{equation}    
\label{eq:descent-lemma}
\mathcal{L}(\mathbf{\bar{w}}^{(k+1)}) \leq \mathcal{L}(\mathbf{\bar{w}}^{(k)}) + \langle \nabla \mathcal{L}(\mathbf{\bar{w}}^{(k)}), \mathbf{\bar{w}}^{(k+1)} - \mathbf{\bar{w}}^{(k)} \rangle + \frac{L}{2} \|\mathbf{\bar{w}}^{(k+1)} - \mathbf{\bar{w}}^{(k)}\|^2.
\end{equation}

For the \textbf{first inner product term}, we have
\begin{subequations}
\begin{align}
&\langle \nabla \mathcal{L}(\bar{\vw}^{(k)}), \bar{\vw}^{(k+1)} - \bar{\vw}^{(k)} \rangle
= \langle T\nabla \mathcal{L}(\bar{\vw}^{(k)}), (\bar{\vw}^{(k,T)} - \bar{\vw}^{(k)})/T 
+ \eta\nabla \mathcal{L}(\bar{\vw}^{(k)}) - \eta\nabla \mathcal{L}(\bar{\vw}^{(k)}) \rangle\\
&= -\eta T \|\nabla \mathcal{L}(\bar{\vw}^{(k)})\|^2 + \left\langle T\nabla \mathcal{L}(\bar{\vw}^{(k)}), \eta \nabla \mathcal{L}(\bar{\vw}^{(k)}) + \frac{\bar{\vw}^{(k,T)} - \bar{\vw}^{(k)}}{T} \right\rangle\\
&= -\eta T \|\nabla \mathcal{L}(\bar{\vw}^{(k)})\|^2 
\! + \! \left\langle T \nabla \mathcal{L}(\bar{\vw}^{(k)}), 
\frac{\eta}{T} \frac{1}{m} \sum_{i=1}^{m} \! \sum_{t=1}^{T} \nabla \mathcal{L}_i(\bar{\vw}^{(k)}) 
\! - \! \frac{\eta}{T} \frac{1}{m} \sum_{i=1}^{m} \! \sum_{t=1}^{T} 
\bigl( \nabla \mathcal{L}_i(\vw_i^{(k,t)}) \! + \! \Delta_i^{\text{NTK}}(\vw_i^{(k,t)}) \bigr) \right\rangle\\
&= -\eta T \|\nabla \mathcal{L}(\bar{\vw}^{(k)})\|^2 
+ \left\langle T\nabla \mathcal{L}(\bar{\vw}^{(k)}), 
\frac{\eta}{T} \frac{1}{m} \sum_{i=1}^{m} \sum_{t=1}^{T} 
\left( \nabla \mathcal{L}_i(\bar{\vw}^{(k)}) - \nabla \mathcal{L}_i (\vw_i^{(k,t)}) - \Delta_i^{\text{NTK}} (\vw_i^{(k,t)}) \right)\right\rangle\\
&\leq -\eta T \|\nabla \mathcal{L}(\bar{\vw}^{(k)})\|^2 
+ \eta \|\nabla \mathcal{L}(\bar{\vw}^{(k)})\| 
\cdot \left\| \frac{1}{m} \sum_{i=1}^{m} \sum_{t=1}^{T} 
\left( \nabla \mathcal{L}_i(\bar{\vw}^{(k)}) - \nabla \mathcal{L}_i (\vw_i^{(k,t)}) \right)\right\| 
+ \eta T \delta_{\text{NTK}} B.
\end{align}
\end{subequations}
Using Assumption \ref{assum:smooth}, we have
\begin{equation}
\langle \nabla \mathcal{L}(\bar{\vw}^{(k)}), \bar{\vw}^{(k+1)} - \bar{\vw}^{(k)} \rangle \leq -\eta T \|\nabla \mathcal{L}(\bar{\vw}^{(k)})\|^2 
+ \eta LT\|\nabla \mathcal{L}(\bar{\vw}^{(k)})\| 
\cdot \frac{1}{m} \sum_{i=1}^{m} \|\bar{\vw}^{(k)} - \vw_i^{(k,t)}\| 
+ \eta T \delta_{\text{NTK}} B.
\end{equation}
Using Young's Inequality $a\leq \frac{a^2}{2\alpha}+\frac{\alpha}{2}$, we can write the norm in terms of the squared norm where we set:
\begin{equation}
    a =  \|\bar{\vw}^{(k)} - \vw_i^{(k, t)}\|,\quad\alpha = 1/L,
\end{equation}
which yields the bound:
\begin{equation}
\eta LT \|\nabla \mathcal{L}(\bar{\vw}^{(k)})\| 
\cdot \frac{1}{m} \sum_{i=1}^{m} \|\bar{\vw}^{(k)} - \vw_i^{(k,t)}\|
\leq 
\frac{\eta T}{2} \|\nabla \mathcal{L}(\bar{\vw}^{(k)})\|^2 
+ \frac{\eta L^2 T}{2} \frac{1}{m} \sum_{i=1}^{m} \|\bar{\vw}^{(k)} - \vw_i^{(k,t)}\|^2.
\end{equation}
With Jensen's inequality and Lemma $\ref{lem:avg-accumulate-difference}$:
\begin{subequations}
\begin{equation}
\eta LT \|\nabla \mathcal{L}(\bar{\vw}^{(k)})\| 
\cdot \frac{1}{m} \sum_{i=1}^{m} \|\bar{\vw}^{(k)} - \vw_i^{(k,t)}\|\leq \frac{\eta T}{2} \|\nabla \mathcal{L}(\bar{\vw}^{(k)})\|^2 
+ \frac{\eta L^2 T}{2} \frac{1}{m} \sum_{i=1}^{m} \frac{1}{m} \sum_{i=1}^{m} \|\vw_i^{(k,t)} - \vw_i^{(k)}\|^2,
\end{equation}
\begin{equation}
\leq \frac{\eta T}{2} \|\nabla \mathcal{L}(\bar{\vw}^{(k)})\|^2 
+ \frac{\eta L^2 T}{2} \left[ 16 \eta^2 T^2 (\delta_{\text{NTK}}^2 + \sigma_g^2 
+ \frac{1}{m} \sum_{i=1}^{m} \|\nabla \mathcal{L}(\vw_i^{(k)})\|^2 \bigr) \right].
\end{equation}
\end{subequations}
Substituting this bound above, the inner product term is bounded by:
\begin{multline}
\langle \nabla \mathcal{L}(\bar{\vw}^{(k)}), \bar{\vw}^{(k+1)} - \bar{\vw}^{(k)} \rangle \leq -\eta T \|\nabla \mathcal{L}(\bar{\vw}^{(k)})\|^2 
+ \frac{\eta T}{2} \|\nabla \mathcal{L}(\bar{\vw}^{(k)})\|^2 
\\+   8 \eta^3 L^2 T^3 \left[ \delta_{\text{NTK}}^2 + \sigma_g^2 
+ \frac{1}{m} \sum_{i=1}^{m} \|\nabla \mathcal{L}(\vw_i^{(k)})\|^2 \right]
+ \eta T \delta_{\text{NTK}} B.
\end{multline}

For the \textbf{second term}, we have 
\begin{equation}
\frac{L}{2} \|\bar{\vw}^{(k+1)} - \bar{\vw}^{(k)}\|^2
= \frac{L}{2} \|\bar{\vw}^{(k,T)} - \bar{\vw}^{(k)}\|^2 \leq \frac{L}{2} \cdot \frac{1}{m} \sum_{i=1}^{m} \|\vw_i^{(k,T)} - \vw_i^{(k)}\|^2. 
\end{equation}
Using results from Lemma \ref{lem:avg-accumulate-difference}
\begin{equation}
\frac{L}{2} \|\bar{\vw}^{(k+1)} - \bar{\vw}^{(k)}\|^2 \leq 8\eta^2  T^2 L  \left[ \delta_{\text{NTK}}^2 + \sigma_g^2 + \frac{1}{m} \sum_{i=1}^{m} \|\nabla \mathcal{L}(\vw_i^{(k)})\|^2 \right].
\end{equation}
From here, we can write  (\ref{eq:descent-lemma}) as
% \begin{multline}    
% \label{eq:descent-lemma}
% \mathcal{L}_i(\mathbf{\bar{w}^{(k+1)}}) - \mathcal{L}_i(\mathbf{\bar{w}^{(k)}})\leq  \\
% -\frac{\eta T}{2} \|\nabla \mathcal{L}(\bar{\vw}^{(k)})\|^2 
% + \frac{\eta L^2 T}{2}\bigl[16\eta^2 T^2  (\delta_{\text{NTK}}^2 + \sigma_g^2) + 16\eta^2 T^2 \frac{1}{m} \sum_{i=1}^{m} \|\nabla \mathcal{L}(\vw_i^{(k)})\|^2\bigr]
% + \eta T \delta_{\text{NTK}} B\\
% + 
% \frac{L}{2}\bigl(16\eta^2 T^2  (\delta_{\text{NTK}}^2 + \sigma_g^2) + 16\eta^2 T^2 \frac{1}{m} \sum_{i=1}^{m} \|\nabla \mathcal{L}(\vw_i^{(k)})\|^2\bigr)
% \end{multline}
\begin{multline}
    \mathcal{L}({\bar{\vw}^{(k+1)}}) - \mathcal{L}({\bar{\vw}^{(k)}})\leq
    -\frac{\eta T}{2} \|\nabla \mathcal{L}(\bar{\vw}^{(k)})\|^2 + 
    8\eta^2 T^2 L({\eta L T+1})(\delta_{\text{NTK}}^2 + \sigma_g^2) + \eta T \delta_{\text{NTK}} B  \\
    % +({8\eta^3T^3L^2}+ 8\eta^2T^2L)
    +8\eta^2T^2L(\eta T L + 1)
    \frac{1}{m}\sum_{i=1}^{m} \|\nabla \mathcal{L}(\vw_i^{(k)})\|^2.
\end{multline}
Substituting Lemma \ref{lem:avg-clients-grad}, we obtain
%\begin{multline}
%    \mathcal{L}({\bar{\vw}^{(k+1)}}) - \mathcal{L}({\bar{\vw}^{(k)}})\leq
%    -\frac{\eta T}{2} \|\nabla \mathcal{L}(\bar{\vw}^{(k)})\|^2 + 
%    8\eta^2 T^2 L({\eta L T+1})  (\delta_{\text{NTK}}^2 + \sigma_g^2) + \eta T \delta_{\text{NTK}} B  \\
%    + 8\eta^2T^2L(\eta T L + 1) \left[\frac{2L^2 C_1 \eta^2}{(1 - \lambda)^2} + 2\|\nabla \mathcal{L}(\bar{\vw}^{(k)})\|^2 \right],
%\end{multline}
\begin{multline}
    \mathcal{L}({\bar{\vw}^{(k+1)}}) - \mathcal{L}({\bar{\vw}^{(k)}})\leq
    - \left[ \tfrac{\eta T}{2} - 16\eta^2T^2L(\eta T L + 1) \right] \|\nabla \mathcal{L}(\bar{\vw}^{(k)})\|^2 + 
    8\eta^2 T^2 L({\eta L T+1}) (\delta_{\text{NTK}}^2 + \sigma_g^2) + \eta T \delta_{\text{NTK}} B  \\
    + 8\eta^2T^2L(\eta T L + 1) \frac{2L^2 C_1 \eta^2}{(1 - \lambda)^2}.
\end{multline}
Taking the sum over $K$ rounds, we obtain
\begin{multline}
    \mathcal{L}({\bar{\vw}^{(K+1)}}) - \mathcal{L}({\bar{\vw}^{(1)}})\leq
    -K \left[ \tfrac{\eta T}{2} - 16\eta^2T^2L(\eta T L + 1) \right] \min_{1\leq k \leq K}{\|\nabla \mathcal{L}(\bar{\vw}^{(k)})\|^2} + 
    K (8\eta^2 T^2 L) ({\eta L T+1}) (\delta_{\text{NTK}}^2 + \sigma_g^2) \\+ K\eta T \delta_{\text{NTK}} B + 
    K(8\eta^2T^2L)(\eta T L + 1) \frac{2L^2 C_1 \eta^2}{(1 - \lambda)^2},
\end{multline}
By moving terms around, we can finally prove the theorem below:
\begin{multline}\label{eq:gradient_norm_bound} \min_{1 \leq k \leq K} \left\| \nabla \mathcal{L}\left( \bar{\vw}^{(k)} \right) \right\|^2 \leq \frac{ \mathcal{L}\left( \bar{\vw}^{(1)} \right) - \mathcal{L}\left( \bar{\vw}^{(K+1)} \right) }{ K \left[ \frac{\eta T}{2} - 16\eta^2T^2L(\eta T L + 1) \right] } \\ + \frac{ 8 \eta^2 T^2 L\left( {\eta L T + 1 } \right)  (\delta_{\text{NTK}}^2 + \sigma_g^2) + \eta T \delta_{\text{NTK}} B +  8\eta^2T^2L(\eta T L + 1)  \cdot \frac{2 L^2 C_1 \eta^2}{(1 - \lambda)^2} }{ \frac{\eta T}{2} - 16\eta^2T^2L(\eta T L + 1) }. 
\end{multline} 
Wrapping this up in constants, we can rewrite the above inequality as
\begin{equation}
\min_{1 \leq k \leq K} \left\| \nabla \mathcal{L}\left( \bar{\vw}^{(k)} \right) \right\|^2  \leq 
\frac{2\bigr[\mathcal{L}( \bar{\vw}^{(1)} ) - \mathcal{L}^*\bigl]}{K\gamma(T,\eta)}+\alpha(\eta,T,\delta_{\text{NTK}})+\beta(\eta, T,\delta_{\text{NTK}},\lambda),
\end{equation}
where 
\begin{subequations}
\begin{align}
    \gamma(T,\eta) &= \eta T - 32\eta^2T^2L(\eta T L + 1), \\
    \alpha(\eta,T,\sigma_g,\delta_{\text{NTK}}) &= \frac{16 \eta^2 T^2 L\left( \eta L T + 1 \right) (\delta_{\text{NTK}}^2 + \sigma_g^2) + 2\eta T \delta_{\text{NTK}} B }{\gamma(T,\eta)}, \\
    \beta(\eta, T, \sigma_g, \delta_{\text{NTK}}, \lambda) &= \frac{512 \eta^4 T^4 L^3(\eta T L + 1) (\delta_{\text{NTK}}^2 + \sigma_g^2 + B^2)}{(1 - \lambda)^2\gamma(T,\eta)}.
\end{align}
\end{subequations}

\end{document}

%% file: math_commands.tex
\usepackage{amsmath,amsfonts,bm}

\def\eqref#1{equation~\ref{#1}}
\def\1{\bm{1}}

\def\vf{{\bm{f}}}

\def\vq{{\bm{q}}}

\def\vv{{\bm{v}}}
\def\vw{{\bm{w}}}
\def\vx{{\bm{x}}}
\def\vy{{\bm{y}}}

\def\mJ{{\bm{J}}}

\def\mR{{\bm{R}}}

\DeclareMathAlphabet{\mathsfit}{\encodingdefault}{\sfdefault}{m}{sl}
\SetMathAlphabet{\mathsfit}{bold}{\encodingdefault}{\sfdefault}{bx}{n}

\newcommand{\R}{\mathbb{R}}

%% file: tex_figures/fig_rounds_fmnist_selected.tex
\begin{figure*}[!t]
\centering
  \begin{subfigure}[b]{0.3515\textwidth}
    \centering
    \includegraphics[width=\textwidth]{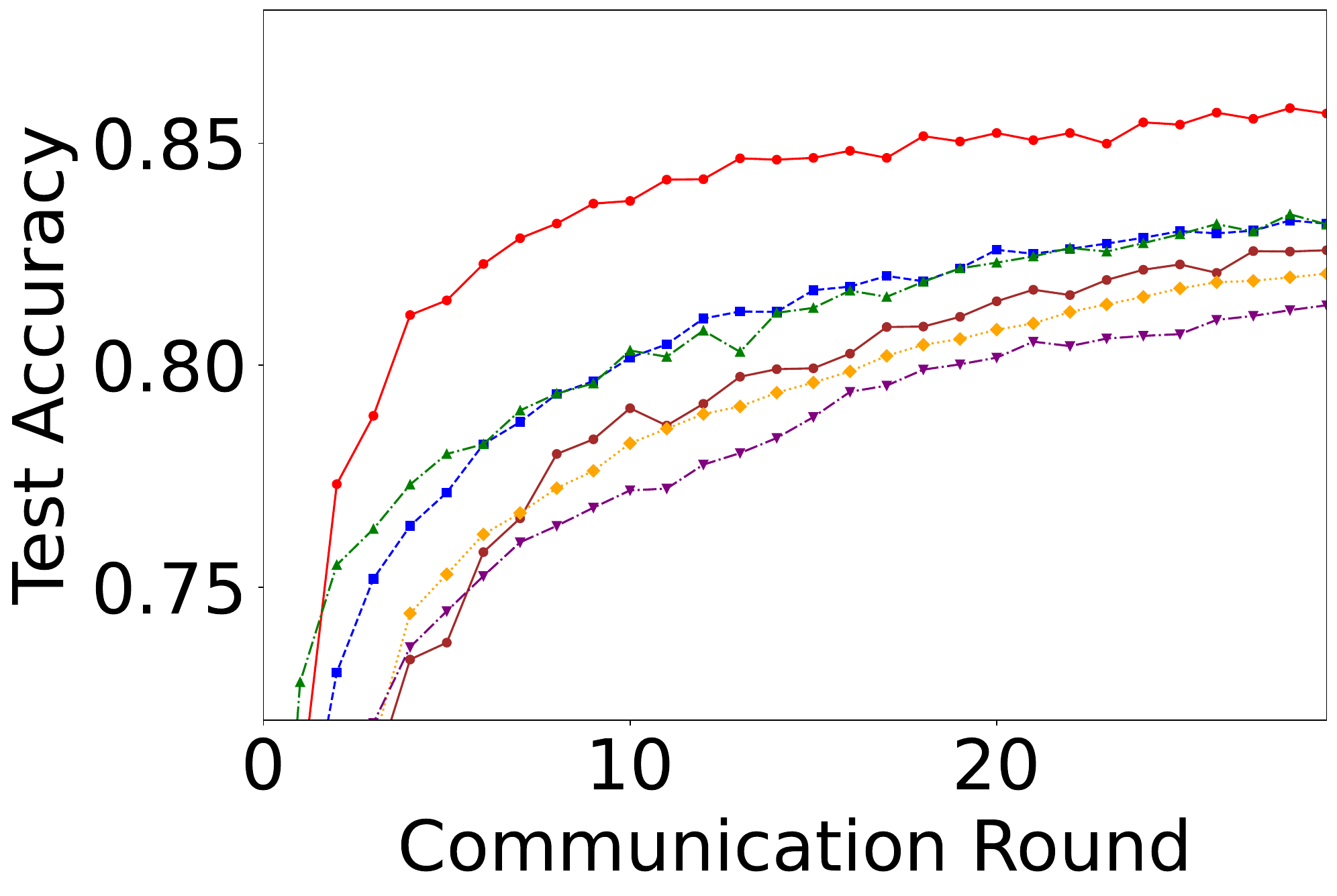}
  \end{subfigure}
  %\begin{subfigure}[b]{0.299\textwidth}
  %  \centering
  %  \includegraphics[width=\textwidth]{images/alpha05_convergence.pdf}
  %  \subcaption{Moderately non-IID}
  %\end{subfigure}  
  \begin{subfigure}[b]{0.29\textwidth}
    \centering
    \includegraphics[width=\textwidth]{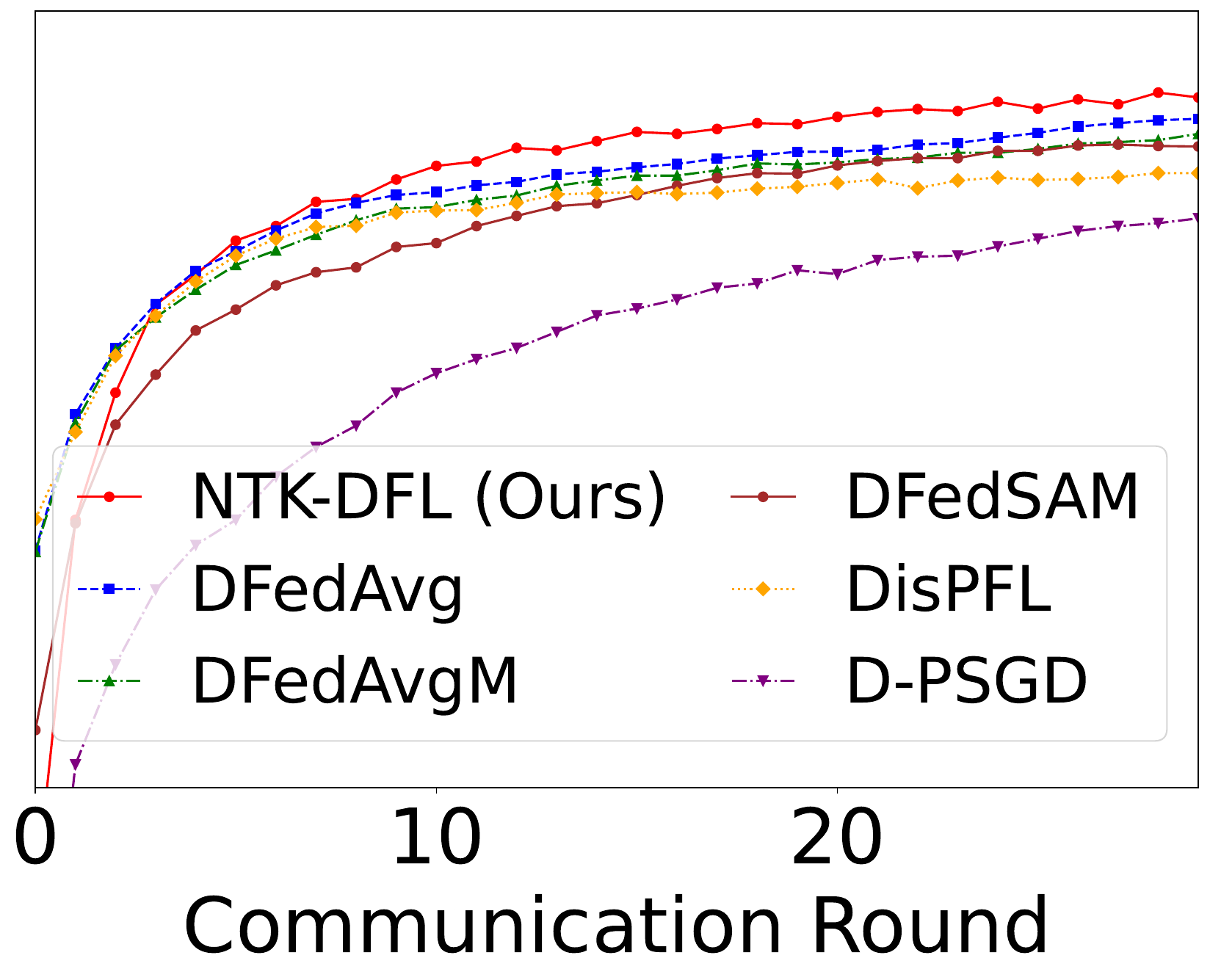}
  \end{subfigure}    
  \begin{subfigure}[b]{0.325\textwidth}
    \centering
    % \caption{Communication rounds for threshold accuracy}
    \label{tab:comm-rounds}
    \include{tex_figures/fig_rounds_table}
    \vspace{0mm}
  \end{subfigure}   
% \captionsetup{margin={1pt,0pt}}
\caption{Convergence of different methods on Fashion-MNIST for (left) highly non-IID with $\alpha=0.1$ and (middle) IID settings. (Right)~The table displays the communication rounds required to reach 85\% test accuracy on Fashion-MNIST. We observe increased improvement in NTK-DFL convergence over baselines for more heterogeneous settings.
% \CommentWong{Refer to Kai's paper's source code, align this figure to the left taking 2/3 of the space, and align the ``Communication rounds'' table to the right 1/3. Since both the plots and table study communication rounds, they can be put together to save space.}
}
\vspace{-2mm}
\label{fig:fashion_mnist_convergence}
\end{figure*}

%% file: tex_figures/fig_rounds_table.tex
\resizebox{\linewidth}{!}{
    \begin{tabular}{@{}lccc@{}}
    \toprule
    \textbf{Method} & \textbf{IID} & \boldmath{$\alpha = 0.5$} & \boldmath{$\alpha = 0.1$} \\
    \midrule
    % NTK-FL & 73 & 85 & 180 \\
    DFedAvg & 18 & 31 & 83 \\
    DFedAvgM & 23 & 43 & 86 \\
    DFedSAM  & 24 & 45 & 200+ \\
    DisPFL & 43 & 87 & 200+ \\
    D-PSGD & 53 & 79 & 125 \\
    \textbf{NTK-DFL} & \textbf{12} & \textbf{17} & \textbf{18} \\
    \bottomrule
    \end{tabular}
}
% \begin{wrapfigure}{r}{0.45\textwidth}
% \centering
% \vspace{-5mm}
% \caption{Communication rounds required to reach 85\% test accuracy on Fashion-MNIST.
% % \CommentWong{Use scalebox to ensure the table's font size is comparable or slightly smaller than the main text size. Just the table width to make a precise alignment. Fix ALL.}
% }
% \label{tab:comm-rounds}
% \resizebox{\linewidth}{!}{
% \begin{tabular}{@{}lccc@{}}
% \toprule
% \textbf{Method} & \textbf{IID} & \boldmath{$\alpha = 0.5$} & \boldmath{$\alpha = 0.1$} \\
% \midrule
% NTK-FL & 73 & 85 & 180 \\
% DFedAvg & 18 & 31 & 83 \\
% DFedAvgM & 23 & 43 & 86 \\
% DisPFL & 43 & 87 & 200+ \\
% D-PSGD & 53 & 79 & 125 \\
% \textbf{NTK-DFL} & \textbf{12} & \textbf{17} & \textbf{18} \\
% \bottomrule
% \end{tabular}
% }
% \end{wrapfigure}
% Tried something other than scalebox to make alginment better
% \begin{wrapfigure}{r}{0.5\textwidth}
% \centering
% \vspace{-10mm}
% \caption{Communication rounds required to reach 85\% test accuracy on Fashion-MNIST.\CommentWong{Use scalebox to ensure the table's font size is comparable or slightly smaller than the main text size. Just the table width to make a precise alignment. Fix ALL.}}
% \label{tab:comm-rounds}
% \scalebox{0.95}{
% \begin{tabular}{lccc}
% \toprule
% \textbf{Method} & \textbf{IID} & \boldmath{$\alpha = 0.5$} & \boldmath{$\alpha = 0.1$} \\
% \midrule
% NTK-FL & 73 & 85 & 180 \\
% DFedAvg & 18 & 31 & 83 \\
% DFedAvgM & 23 & 43 & 86 \\
% DisPFL & 43 & 87 & 200+ \\
% D-PSGD & 53 & 79 & 125 \\
% \textbf{NTK-DFL} & \textbf{12} & \textbf{17} & \textbf{18} \\
% \bottomrule
% \end{tabular}
% }
% \end{wrapfigure}

%% file: tex_figures/fig_neighbor_hetero.tex
\begin{figure*}[!t]
    % \centering
    \includegraphics[width=0.53\linewidth]{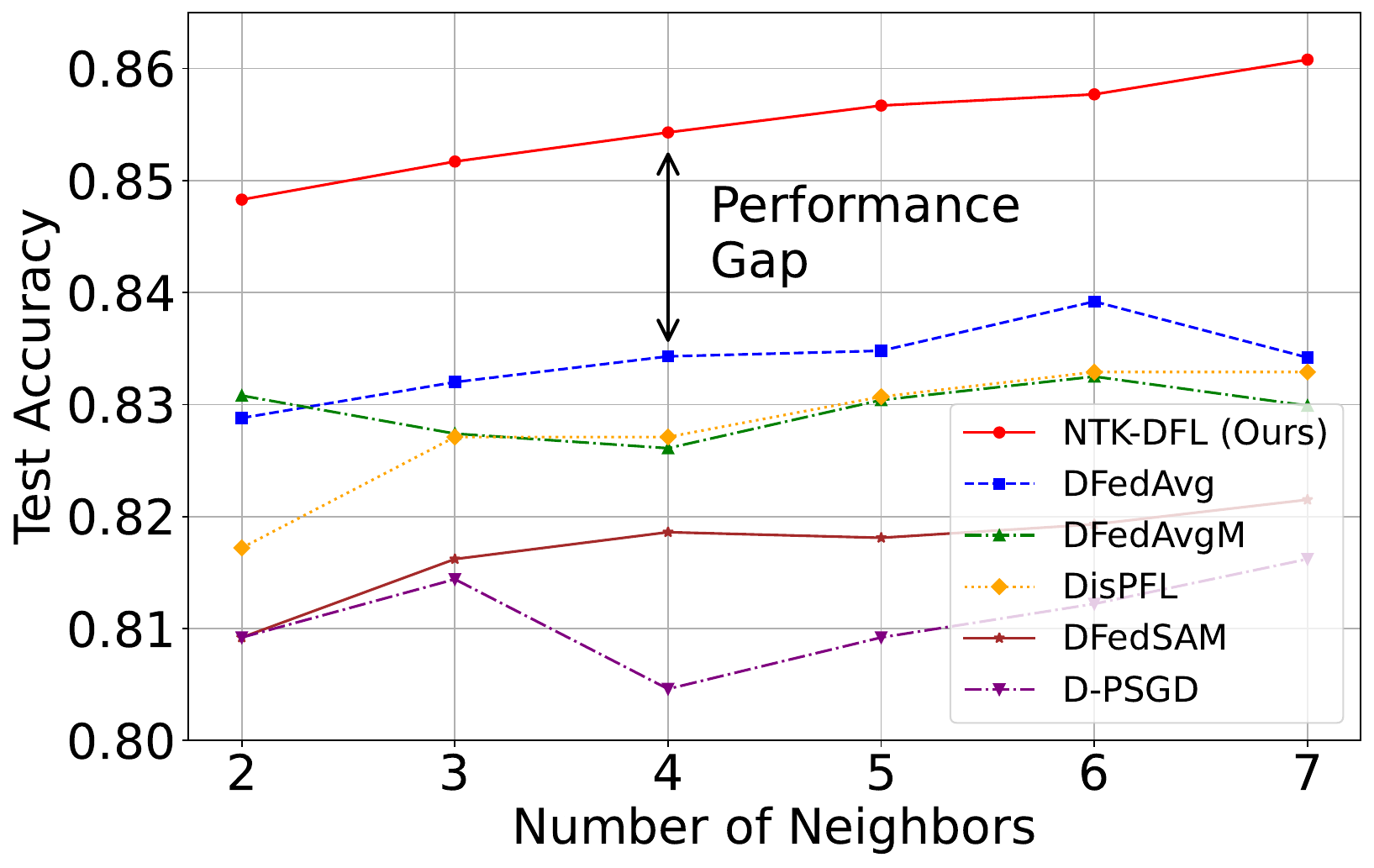}
    \includegraphics[width=0.463\linewidth]{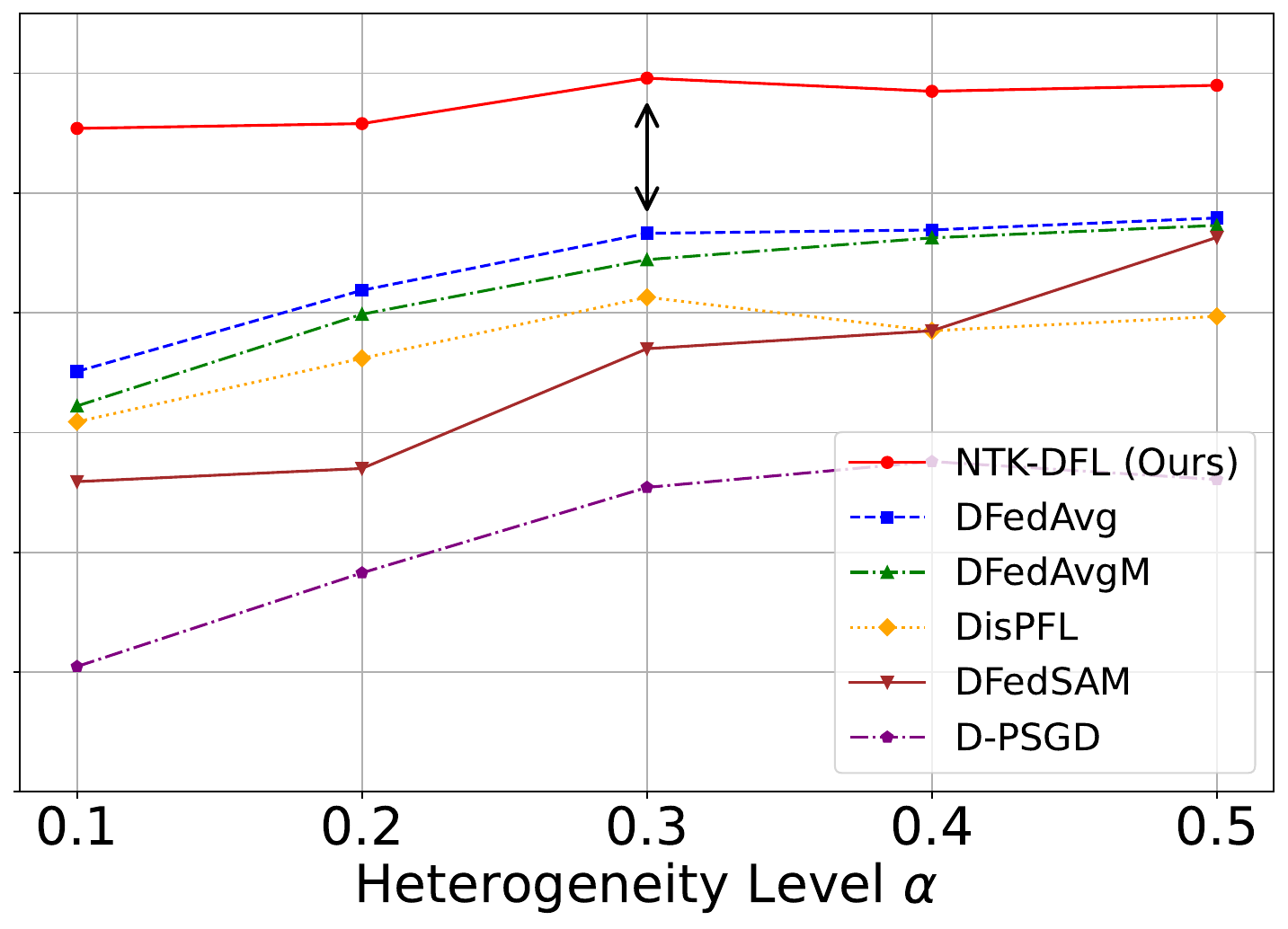}
    \caption{Performance of NTK-DFL vs. (left) sparsity level and (right) heterogeneity level (smaller $\alpha$ $\to$ more heterogeneous). NTK-DFL outperforms the baselines and the gains are stable as the factors vary. 
    % \CommentWong{The large and stable performance gap should be one of the major merit points of this paper. Please incorporate it into the end of Section 1.}
    }
    \vspace{-2mm}
    \label{fig:topology_compare}
\end{figure*}

%% file: tex_figures/fig_heterogeneity_compare.tex
\begin{figure}
    \centering
    % \vspace{-6pt}
    {\includegraphics[width=1\linewidth]{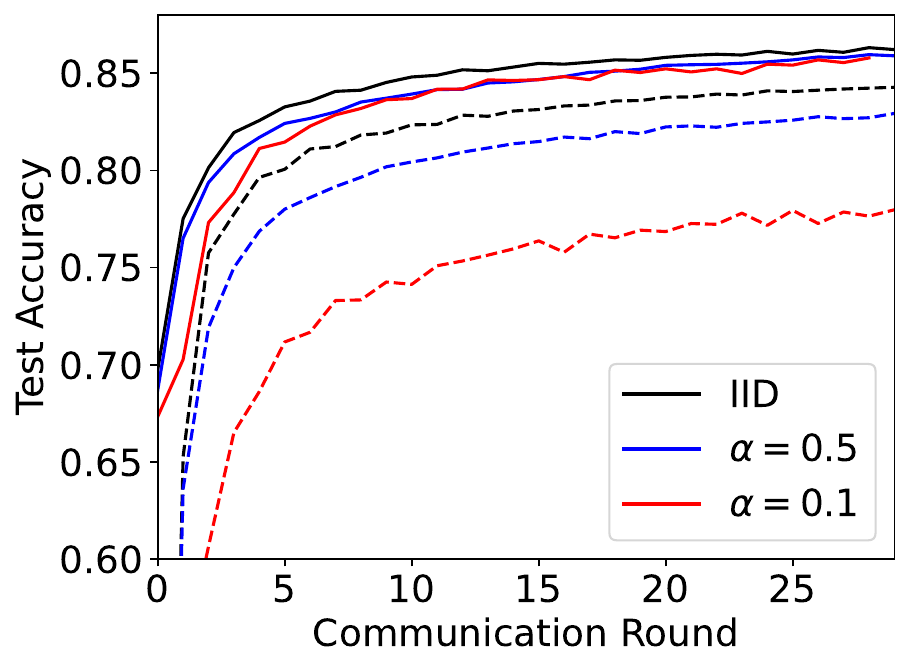}\vspace{-4mm}}
    \captionsetup{width=0.45\textwidth, margin={3pt,0pt}}
    \caption{Performance gains of model averaging on convergence, trained on Fashion-MNIST. Solid lines correspond to the accuracy of the aggregated global model, whereas dotted lines correspond to the mean accuracy across client models. 
    NTK-DFL's aggregated model maintains high performance, whereas mean client accuracy declines significantly with increased heterogeneity.}
    \label{fig:heterogeneity_compare}
\end{figure}

%% file: tex_math/lemma1.tex
\begin{lemma}
\label{lem:mixing-projection} [Following \citet{d-psgd}]
For any \( k \in \mathbb{Z}^{+} \), the mixing matrix \( \mathbf{M} \in \mathbb{R}^{m \times m} \) satisfies the inequality 
\begin{equation}
\| \mathbf{M}^k - \mathbf{P} \|_{\text{op}} \leq \lambda^k,   
\end{equation}
where the parameter \( \lambda \) is defined as 
\begin{equation}
\lambda := \max\{ |\lambda_2(\mathbf{M})|, |\lambda_m(\mathbf{M})| \}.
\end{equation}
Here, the spectral norm of a matrix \( \mathbf{A} \) is denoted by \( \| \mathbf{A} \|_{\text{op}} \), and the matrix \( \mathbf{P} \) is given by
\begin{equation}
\mathbf{P} = \frac{\mathbbm{1} \mathbbm{1}^\top}{m} \in \mathbb{R}^{m \times m},
\end{equation}
where \( \mathbbm{1} \) is the all-one vector \( [1, 1, \dots, 1]^\top \in \mathbb{R}^{m} \).
\end{lemma}

%% file: tex_math/lemma2.tex
\begin{lemma}
\label{lem:per-step-difference}
Assume that Assumptions \ref{assum:bounded-gradient} and \ref{assum:ntk-error} hold. Let \( \vw^{(k,t)} \) denote the model weights at round \( k \) and timestep \( t \), and let the weight evolution be governed by the NTK-DFL approach. Then, it follows that
\begin{equation}
\|\vw^{(k,t+1)} - \vw^{(k,t)}\|^2 \leq \eta^2(B+\delta_{\text{NTK}})^2,
\end{equation}
for \( 1 \leq k \leq K \).
\end{lemma}

%% file: tex_math/lemma2_proof.tex
\textbf{Proof}. [Following \citet{dfedavg}] \\
We begin with the NTK update step
\begin{equation}
\vw_i^{(k,t+1)} - \vw_i^{(k,t)} = \frac{\eta}{N_i^{(k)}} \nabla f_i(\mathbf{X}_i; \vw_i^{(k,0)})^\top \mathbf{r}_i(\mathbf{X}_i, \mathbf{Y}_i, \vw_i^{(k,t)}),
\end{equation}
% where $ r_i(\vw, {x}_i, {y}_i) = \nabla_f \ell(f_i({x}_i; \vw), {y}_i)$ is the (possibly transformed) residual over the data between client $i$ and its neighbors $j \in \mathcal{N}_i$. 
where \( \mathbf{r}_i(\mathbf{X}_i, \mathbf{Y}_i, \vw) = \nabla_f \ell(f_i(\mathbf{X}_i; \vw), \mathbf{Y}_i) \) is the (possibly transformed) residual over the data between client \( i \) and its neighbors \( j \in \mathcal{N}_i \).  
The specific \( \mathbf{r}_i \) depends on the loss of choice, and we provide general derivation that is loss-agnostic. Define the NTK-approximated gradient of the local loss function \( \mathcal{L}_i \) with respect to the model weights \( \vw_i \), using the Jacobian evaluated at \( t = 0 \), we obtain
\begin{equation}
\vw_i^{(k,t+1)} - \vw_i^{(k,t)} = \eta \left( \nabla \mathcal{L}_i(\vw_i^{(k,t)}) + \Delta_i^{\text{NTK}}(\vw_i^{(k,t)}) \right),
\end{equation}
where
\begin{subequations}
\begin{align}
\nabla_{\vw} \mathcal{L}_i^{\text{NTK}} (\vw_i^{(k,t)}) 
&= \frac{1}{\tilde{N}^{(k)}_i} \nabla_{\vw} f(\tilde{\mathbf{X}}_i, \vw_i^{(k,0)})^\top \mathbf{r}_i(\tilde{\mathbf{X}}_i, \tilde{\mathbf{y}}_i, \vw_i^{(k,t)}), \\
\Delta_i^{\text{NTK}} 
&= \nabla_{\vw} \mathcal{L}_i(\vw^{(k,t)}) - \nabla_{\vw} \mathcal{L}_i^{\text{NTK}}(\vw^{(k,t)}).
\end{align}
\end{subequations}
% \begin{equation}
% \nabla_{\vw} \mathcal{L}_i^{\text{NTK}} (\vw_i^{(k,t)}) 
% = \frac{1}{\tilde{N}^{(k)}_i} \nabla_{\vw} f(\tilde{\mathbf{X}}_i, \vw_i^{(k,0)})^\top \mathbf{r}_i(\tilde{\mathbf{X}}_i, \tilde{\mathbf{y}}_i, \vw_i^{(k,t)}), \quad  
% \Delta_i^{\text{NTK}} = \nabla_{\vw} \mathcal{L}_i(\vw^{(k,t)}) - \nabla_{\vw} \mathcal{L}_i^{\text{NTK}}(\vw^{(k,t)}),
% \end{equation}
Using Assumptions \ref{assum:bounded-gradient} and \ref{assum:ntk-error} and taking the norm:
\begin{equation}
\|\vw_i^{(k,t+1)} - \vw_i^{(k,t)}\| \leq \eta(B + \delta_{\text{NTK}}),
\end{equation}
therefore completing the proof.\\

%% file: tex_math/lemma3.tex
\begin{lemma}
\label{lem:avg-accumulate-difference} 
Given the stepsize \( 0 < \eta \leq \frac{1}{8LT} \), assume \( \vw_i^{(k,t)} \) and \( \vw_i^{(k)} \) are generated by the decentralized NTK-FL algorithm for all \( i \in \{1,2,\dots,m\} \). If Assumptions \ref{assum:smooth}, \ref{assum:bounded variance}, and \ref{assum:ntk-error} hold, it follows that
\begin{equation}
\frac{1}{m} \sum_{i=1}^{m} \|\vw_i^{(k,t)} - \vw_i^{(k,0)}\|^2 
\leq 
16\eta^2 T^2 \left[ \delta_{\text{NTK}}^2 + \sigma_g^2 + \tfrac{1}{m} \sum_{i=1}^{m} \|\nabla \mathcal{L}(\vw_i^{(k)})\|^2 \right],
\end{equation}
for \( 1 \leq t \leq K \).
\end{lemma}

\textbf{Proof}. [Following \citet{dfedavg}]\\
Decomposing the difference into global variance, the NTK difference, and the difference between $t=0$ and $t$,
\begin{multline}
\vw_i^{(k,t+1)} - \vw_i^{(k,0)}= \vw_i^{(k,t)} - \vw_i^{(k,0)} 
- \eta \Big[ \Delta_i^{\text{NTK}}(\vw_i^{(k,t)})+ \left( \nabla \mathcal{L}_i(\vw_i^{(k,t)}) - \nabla \mathcal{L}_i(\vw_i^{(k,0)}) \right) \\\quad + \left( \nabla \mathcal{L}_i(\vw_i^{(k,0)}) - \nabla \mathcal{L}(\vw^{(k,0)}) \right) + \nabla \mathcal{L}(\vw_i^{(k,0)}) \Big].
\end{multline}
We consider two terms such that \(\|\vw_i^{(k,t+1)} - \vw_i^{(k,0)}\|^2\leq  \mathrm{I} + \mathrm{II} \), using the Cauchy inequality and Assumptions \ref{assum:bounded variance} and \ref{assum:ntk-error}
\begin{multline}
\mathrm{I} \leq ( 1 + \tfrac{1}{2T-1}) \left\| \vw_i^{(k,t)} - \vw_i^{(k,0)} \right\|^2,\ 
\mathrm{II} \leq  8T\eta^2\Bigl( \delta_{\text{NTK}}^2  
+ \|\nabla \mathcal{L}_i(\vw_i^{(k,t)})\ - \nabla \mathcal{L}_i(\vw_i^{(k,0)})\|^2 
  + \sigma_g^2
+ \|\nabla \mathcal{L}(\vw_i^{(k,0)})\|^2\bigr).
\end{multline}
From here, rewriting term $\mathrm{II}$ using Assumption \ref{assum:smooth}, we have
\begin{multline}
     \left\| \vw_i^{(k,t+1)} - \vw_i^{(k,0)} \right\|^2\leq 
     ( 1 + \tfrac{1}{2T-1}+8T\eta^2L^2) \left\| \vw_i^{(k,t)} - \vw_i^{(k,0)} \right\|^2 + 8T\eta^2(\delta_{\text{NTK}}^2+\sigma_g^2)+8T\eta^2\|\nabla \mathcal{L}(\vw_i^{(k,0)})\|^2.
\end{multline}
Given an appropriately small step size, we simplify $ 1 + \frac{1}{2T-1}+8T\eta^2L^2 \leq 1 + \frac{1}{T-1}$. Solving the recursive relationship yields 
\begin{subequations}
\begin{align}
    \frac{1}{m}\sum_{i=1}^m\left\| \vw_i^{(k,t)} - \vw_i^{(k,0)} \right\|^2 &\leq
    \frac{1}{m}\sum_{i=1}^m\sum_{u=1}^T ( 1 + \tfrac{1}{T-1})^u (8T\eta^2)\bigl[(\delta_{\text{NTK}}^2+\sigma_g^2)+\|\nabla \mathcal{L}(\vw_i^{(k,0)})\|^2\bigr]\\
&\leq 16T^2\eta^2\Big[ \delta_{\text{NTK}}^2+\sigma_g^2 + \tfrac{1}{m}\sum_{i=1}^m\|\nabla \mathcal{L}(\vw_i^{(k,0)})\|^2\Big],
\end{align}
\end{subequations}
with the recursive sum $\sum_{u=1}^T ( 1 + \frac{1}{T-1})^u\leq 2T$ for $T\geq10$. Thus, we have completed the proof. \\

%% file: tex_math/lemma4.tex
\begin{lemma}
\label{lem:dev_from_avg}
Suppose the stepsize satisfies \( 0 < \eta \leq \frac{1}{8LT} \), and let \( \vw_i^{(k,0)} \)  be the model weights produced by the NTK-DFL algorithm for all clients \( i \in \{1,2,\dots,m\} \). Under Assumptions \ref{assum:bounded variance} and \ref{assum:ntk-error} the following bound holds:
\begin{equation}
\frac{1}{m} \sum_{i=1}^{m} \|\vw_i^{(k,0)} - \bar{\vw}^{(k,0)}\|^2 
\leq \frac{16T^2 \eta^2 \left[ \delta_{\text{NTK}} + \sigma_g^2 + \frac{1}{m} \sum_{i=1}^{m} \|\nabla \mathcal{L}(\vw_i^{(k,0)})\|^2 \right] }{(1 - \lambda)^2},
\end{equation}
for all \( 1 \leq k \leq K \).
\end{lemma}

\textbf{Proof.} [Following \citet{dfedavg} and \citet{dfedsam}] \\
With a slight abuse of notation, $\mathbf{Y}^{k} \coloneq [\vw_1^{(k,T)},\vw_2^{(k,T)}, \ldots,\vw_m^{(k,T)}]$ and  $\mathbf{X}^{k} \coloneq [\vw_1^{(k,0)},\vw_2^{(k,0)}, \ldots,\vw_m^{(k,0)}]$, we may write the mixing process as
\begin{equation}
\mathbf{X}^{k+1} = \mathbf{M} \mathbf{Y}^{k} = \mathbf{M} \mathbf{X}^{k} - \boldsymbol{\zeta}^{k},
\end{equation}
where 
\begin{equation}
\boldsymbol{\zeta}^{k} := \mathbf{M}( \mathbf{X}^{k} - \mathbf{Y}^{k}).    
\end{equation}
We note that the order of averaging and updating weights does not affect this proof, as it relies on the spectral gap of the (possibly time-varying) mixing matrix, which does not change over time. We can write recursively
\begin{equation}
\label{eq:mixing-update}
\mathbf{X}^{k} = \mathbf{M}^{k} \mathbf{X}^{0} - \sum_{j=0}^{k-1} \mathbf{M}^{k-1-j} \boldsymbol{\zeta}^{j}.
\end{equation}
We note the following property of the mixing operation:
\begin{equation}
\label{eq:projection-mixing}
    \mathbf{PM} = \mathbf{MP} = \mathbf{P}.
\end{equation} 
From Lemma \ref{lem:mixing-projection} we have that $\|\mathbf{M}^k-\mathbf{P}\|_{\text{op}}\leq\lambda^k$. Left-multiplying both sides of (\ref{eq:mixing-update}) and using the mixing property (\ref{eq:projection-mixing}), we obtain
\begin{equation}
\mathbf{PX}^k=\mathbf{PX}^0-\sum_{j=0}^{k-1}\mathbf{P}\boldsymbol{\zeta}^j=-\sum_{j=0}^{k-1}\mathbf{P}\boldsymbol{\zeta}^j,
\end{equation}
with the initialization $\mathbf{X}^0=\mathbf{0}$. Then, we are led to
\begin{multline}
\|\mathbf{X}^k - \mathbf{P}\mathbf{X}^k\| = \left\| \sum_{j=0}^{k-1} (\mathbf{P} - \mathbf{M}^{k-1-j}) \boldsymbol{\zeta}^j \right\| 
\leq \sum_{j=0}^{k-1} \|\mathbf{P} - \mathbf{M}^{k-1-j}\|_{\mathrm{op}} \|\boldsymbol{\zeta}^j\| 
\leq \sum_{j=0}^{k-1} \lambda^{k-1-j} \|\boldsymbol{\zeta}^j\|.
\end{multline}
With the Cauchy inequality, we obtain
\begin{multline}
\|\mathbf{X}^k - \mathbf{P}\mathbf{X}^k\|^2 \leq \left( \sum_{j=0}^{k-1} \lambda^{\frac{k-1-j}{2}} \cdot \lambda^{\frac{k-1-j}{2}} \|\boldsymbol{\zeta}^j\| \right)^2 
\leq \left( \sum_{j=0}^{k-1} \lambda^{k-1-j} \right) \left( \sum_{j=0}^{k-1} \lambda^{k-1-j} \|\boldsymbol{\zeta}^j\|^2 \right),
\end{multline}
and we can compute that
\begin{equation}
\|\boldsymbol{\zeta}^j\|^2 \leq \|\mathbf{M}\|^2 \cdot \|\mathbf{X}^j - \mathbf{Y}^j\|^2 \leq \|\mathbf{X}^j - \mathbf{Y}^j\|^2.
\end{equation}
Applying Lemma \ref{lem:avg-accumulate-difference}, for all $j$,
\begin{equation}
\frac{1}{m}\|\mathbf{X}^j-\mathbf{Y}^j\|^2\leq 16T^2 \eta^2
\left[ \delta_{\text{NTK}}^2 + \sigma_g^2 + \frac{1}{m} \sum_{i=1}^{m} \|\nabla \mathcal{L}(\vw_i^{(k)})\|^2
\right].
\end{equation}
Finally, we complete the proof with
\begin{equation}
    \frac{1}{m}\|\mathbf{X}^t-\mathbf{PX}^t\|^2\leq \frac{16T^2 \eta^2 \big[ \delta_{\text{NTK}}^2 + \sigma_g^2 +  \frac{1}{m} \sum_{i=1}^{m} \|\nabla \mathcal{L}(\vw_i^{(k)})\|^2 \big]}{(1-\lambda)^2}.
\end{equation}

%% file: tex_math/lemma5.tex
\begin{lemma}
\label{lem:avg-clients-grad}
Using the result from Lemma \ref{lem:dev_from_avg}, we obtain the following bound for the average squared gradient norm:
\begin{equation}
\frac{1}{m} \sum_{i=1}^{m} \|\nabla \mathcal{L}(\vw_i^{(k)})\|^2 
\leq \frac{2L^2 C_1 \eta^2}{(1 - \lambda)^2} + 2\|\nabla \mathcal{L}(\bar{\vw}^{(k)})\|^2,
\end{equation}
% where $C_2 = 16 T^2  (\delta_{\text{NTK}} + \sigma_g) + 16 T^2 \frac{1}{m} \sum_{i=1}^{m} \|\nabla \mathcal{L}(\vw_i^{(k)})\|^2$.
where $C_1 = 16 T^2  (\delta_{\text{NTK}}^2 + \sigma_g^2 + B^2)$.
\end{lemma}

%% file: tex_math/lemma5_proof.tex
\textbf{Proof}.
\\We expand the expression below by adding a difference term:
\begin{equation}
\frac{1}{m} \sum_{i=1}^{m} \|\nabla \mathcal{L}(\vw_i^{(k)})\|^2 
\leq \frac{1}{m} \sum_{i=1}^{m} \|\nabla \mathcal{L}(\vw_i^{(k)}) - \nabla \mathcal{L}(\bar{\vw}^{(k)}) + \nabla \mathcal{L}(\bar{\vw}^{(k)})\|^2.
\end{equation}
Using $\|a+b\|^2\leq2\|a\|^2+2\|b\|^2$, we have
\begin{equation}
\frac{1}{m} \sum_{i=1}^{m} \|\nabla \mathcal{L}(\vw_i^{(k)})\|^2 
\leq \frac{1}{m} \sum_{i=1}^{m} 2\|\nabla \mathcal{L}(\vw_i^{(k)}) - \nabla \mathcal{L}(\bar{\vw}^{(k)})\|^2 + 2\|\nabla \mathcal{L}(\bar{\vw}^{(k)})\|^2.
\end{equation}
With Assumption \ref{assum:smooth}, we can bound the gradient difference as
\begin{equation}
\frac{1}{m} \sum_{i=1}^{m} \|\nabla \mathcal{L}(\vw_i^{(k)})\|^2 \leq 2L^2 \frac{1}{m} \sum_{i=1}^{m} \|\vw_i^{(k)} - \bar{\vw}^{(k)}\|^2 + 2\|\nabla \mathcal{L}(\bar{\vw}^{(k)})\|^2.
\end{equation}
Finally, we substitute the result from Lemma \ref{lem:dev_from_avg} and use Assumption \ref{assum:bounded-gradient} to finish the proof:
\begin{equation}
\frac{1}{m} \sum_{i=1}^{m} \|\nabla \mathcal{L}(\vw_i^{(k)})\|^2 \leq \frac{32L^2 T^2 \eta^2 \bigl(\delta_{\text{NTK}}^2 + \sigma_g^2 + B^2\bigr) }{(1-\lambda)^2} + 2\|\nabla \mathcal{L}(\bar{\vw}^{(k)})\|^2.
\end{equation}

%% file: tex_math/lemma6.tex
\begin{lemma}
\label{lem:delta_ntk_bound}
Assuming an appropriately small step size (such as is chosen in Theorem \ref{thm:main}), Lipschitz continuity for  $\nabla f_i(\mathbf{X}_i;\vw^{(k,t)})$ and a bounded mean residual norm across all clients, i.e. $\frac{1}{N_i}\sum\|\mathbf{r_i}(\mathbf{X}_i, \mathbf{Y}_i;\vw)\|\leq r$, we can bound the error term in Assumption \ref{assum:ntk-error} as follows: 
\begin{equation}
    \delta_{\text{NTK}}^2 \leq16\eta^2T^2L^2r^2(\sigma_g^2+B^2).
\end{equation}
\end{lemma}
\textbf{Proof. }\\
We can express $\Delta_{\text{NTK}}$ as follows
\begin{subequations}
\begin{align}
    \Delta_i^{\text{NTK}} &= \nabla_{\vw} \mathcal{L}_i(\vw^{(k,t)}_i) - \nabla_{\vw} \mathcal{L}_i^{\text{NTK}}(\vw^{(k,t)}_i) \\
    &= \frac{1}{{N}^{(k)}_i} \bigl(\nabla_{\vw} f({\mathbf{X}}_i, \vw_i^{(k,t)})^\top - \nabla_{\vw} f({\mathbf{X}}_i, \vw_i^{(k,0)})^\top\bigr) \mathbf{r}_i({\mathbf{X}}_i, {\mathbf{y}}_i, \vw_i^{(k,t)}).
\end{align}
\end{subequations}
Using the Lipschitz assumption listed above, where the Lipschitz constant for $f_i$ is $L$, we have
\begin{equation}
    \delta_{\text{NTK}}^2 \leq L^2\|\vw_i^{(k,t)}-\vw_i^{(k,0)}\|^2r^2.
\end{equation}
Now, substituting the result from \ref{lem:avg-accumulate-difference}
\begin{equation}
     \delta_{\text{NTK}}^2 \leq L^2 \left[ 16\eta^2 T^2  (\delta_{\text{NTK}}^2 + \sigma_g^2) + 16\eta^2 T^2 B^2 \right]r^2.
\end{equation}
Solving for $\delta_{\text{NTK}}^2$ 
\begin{equation}
    \delta_{\text{NTK}}^2 \leq \frac{16\eta^2T^2r^2L^2(\sigma_g^2+B^2)}{1-16\eta^2T^2r^2}\approx
     16\eta^2T^2r^2L^2(\sigma_g^2+B^2).
\end{equation}
where the second inequality comes from assuming a step size $\eta$ such that, for a large communication round $K$, $16\eta^2T^2r^2 \ll1$. Thus, the proof is complete.

%% file: main.bbl
\begin{thebibliography}{38}
\providecommand{\natexlab}[1]{#1}
\providecommand{\url}[1]{\texttt{#1}}
\expandafter\ifx\csname urlstyle\endcsname\relax
  \providecommand{\doi}[1]{doi: #1}\else
  \providecommand{\doi}{doi: \begingroup \urlstyle{rm}\Url}\fi

\bibitem[Alemohammad et~al.(2021)Alemohammad, Wang, Balestriero, and Baraniuk]{ntk_rnn}
Alemohammad, S., Wang, Z., Balestriero, R., and Baraniuk, R.
\newblock The recurrent neural tangent kernel.
\newblock In \emph{International Conference on Learning Representations}, 2021.

\bibitem[Alistarh et~al.(2018)Alistarh, Hoefler, Johansson, Konstantinov, Khirirat, and Renggli]{topk}
Alistarh, D., Hoefler, T., Johansson, M., Konstantinov, N., Khirirat, S., and Renggli, C.
\newblock The convergence of sparsified gradient methods.
\newblock In \emph{Advances in Neural Information Processing Systems}, volume~31, 2018.

\bibitem[Arora et~al.(2019)Arora, Du, Hu, Li, Salakhutdinov, and Wang]{ntk_cnn}
Arora, S., Du, S.~S., Hu, W., Li, Z., Salakhutdinov, R., and Wang, R.
\newblock On exact computation with an infinitely wide neural net.
\newblock In \emph{Advances in Neural Information Processing Systems}, 2019.

\bibitem[Caldas et~al.(2019)Caldas, Duddu, Wu, Li, Konečný, McMahan, Smith, and Talwalkar]{leaf}
Caldas, S., Duddu, S. M.~K., Wu, P., Li, T., Konečný, J., McMahan, H.~B., Smith, V., and Talwalkar, A.
\newblock {LEAF}: A benchmark for federated settings.
\newblock In \emph{Advances in Neural Information Processing Systems}, 2019.

\bibitem[Dai et~al.(2022)Dai, Shen, He, Tian, and Tao]{dispfl}
Dai, R., Shen, L., He, F., Tian, X., and Tao, D.
\newblock {DisPFL}: Towards communication-efficient personalized federated learning via decentralized sparse training.
\newblock In \emph{International Conference on Machine Learning}, 2022.

\bibitem[Foret et~al.(2021)Foret, Kleiner, Mobahi, and Neyshabur]{sam}
Foret, P., Kleiner, A., Mobahi, H., and Neyshabur, B.
\newblock Sharpness-aware minimization for efficiently improving generalization.
\newblock In \emph{International Conference on Learning Representations}, 2021.

\bibitem[Golikov et~al.(2022)Golikov, Pokonechnyy, and Korviakov]{ntk_survey}
Golikov, E., Pokonechnyy, E., and Korviakov, V.
\newblock Neural tangent kernel: A survey, 2022.

\bibitem[Good(1976)]{dirichlet}
Good, I.~J.
\newblock On the application of symmetric {D}irichlet distributions and their mixtures to contingency tables.
\newblock \emph{The Annals of Statistics}, 4\penalty0 (6), 1976.

\bibitem[Guo et~al.(2025)Guo, Wang, Zhou, Jiang, and Patel]{cross_silo}
Guo, P., Wang, P., Zhou, J., Jiang, S., and Patel, V.~M.
\newblock Enhancing {MRI} reconstruction with cross-silo federated learning.
\newblock In Li, X., Xu, Z., and Fu, H. (eds.), \emph{Federated Learning for Medical Imaging}, The MICCAI Society book Series, pp.\  155--171. Academic Press, 2025.

\bibitem[He et~al.(2016)He, Zhang, Ren, and Sun]{resnet}
He, K., Zhang, X., Ren, S., and Sun, J.
\newblock Deep residual learning for image recognition.
\newblock In \emph{IEEE Conference on Computer Vision and Pattern Recognition}, pp.\  770--778, 2016.

\bibitem[Huang et~al.(2021)Huang, Li, Song, and Yang]{huang2021fl}
Huang, B., Li, X., Song, Z., and Yang, X.
\newblock {FL-NTK}: A neural tangent kernel-based framework for federated learning analysis.
\newblock In \emph{International Conference on Machine Learning}, pp.\  4423--4434. PMLR, 2021.

\bibitem[Jacot et~al.(2018)Jacot, Gabriel, and Hongler]{ntk_jacot}
Jacot, A., Gabriel, F., and Hongler, C.
\newblock Neural tangent kernel: Convergence and generalization in neural networks.
\newblock In \emph{Advances in Neural Information Processing Systems}, pp.\  8580–8589, Red Hook, NY, USA, 2018.

\bibitem[Kairouz et~al.(2021)Kairouz, McMahan, Avent, Bellet, Bennis, Bhagoji, Bonawitz, Charles, Cormode, Cummings, et~al.]{fed_open_problems}
Kairouz, P., McMahan, H.~B., Avent, B., Bellet, A., Bennis, M., Bhagoji, A.~N., Bonawitz, K., Charles, Z., Cormode, G., Cummings, R., et~al.
\newblock Advances and open problems in federated learning.
\newblock \emph{Foundations and Trends{\textregistered} in Machine Learning}, 14:\penalty0 1--210, 2021.

\bibitem[Karimireddy et~al.(2020)Karimireddy, Kale, Mohri, Reddi, Stich, and Suresh]{karimireddy2020scaffold}
Karimireddy, S.~P., Kale, S., Mohri, M., Reddi, S., Stich, S., and Suresh, A.~T.
\newblock Scaffold: Stochastic controlled averaging for federated learning.
\newblock In \emph{International Conference on Machine Learning}, pp.\  5132--5143. PMLR, 2020.

\bibitem[Lecun et~al.(1998)Lecun, Bottou, Bengio, and Haffner]{mnist}
Lecun, Y., Bottou, L., Bengio, Y., and Haffner, P.
\newblock Gradient-based learning applied to document recognition.
\newblock \emph{Proceedings of the IEEE}, 86\penalty0 (11):\penalty0 2278--2324, 1998.

\bibitem[Li et~al.(2020{\natexlab{a}})Li, Sahu, Zaheer, Sanjabi, Talwalkar, and Smith]{fedprox}
Li, T., Sahu, A.~K., Zaheer, M., Sanjabi, M., Talwalkar, A., and Smith, V.
\newblock Federated optimization in heterogeneous networks.
\newblock In \emph{3rd MLSys Conference}, 2020{\natexlab{a}}.

\bibitem[Li et~al.(2020{\natexlab{b}})Li, Sahu, Zaheer, Sanjabi, Talwalkar, and Smith]{li2020federated}
Li, T., Sahu, A.~K., Zaheer, M., Sanjabi, M., Talwalkar, A., and Smith, V.
\newblock Federated optimization in heterogeneous networks.
\newblock \emph{Proceedings of Machine Learning and Systems}, 2:\penalty0 429--450, 2020{\natexlab{b}}.

\bibitem[Lian et~al.(2017)Lian, Zhang, Zhang, Hsieh, Zhang, and Liu]{d-psgd}
Lian, X., Zhang, C., Zhang, H., Hsieh, C.-J., Zhang, W., and Liu, J.
\newblock Can decentralized algorithms outperform centralized algorithms? {A} case study for decentralized parallel stochastic gradient descent.
\newblock In \emph{Advances in Neural Information Processing Systems}, volume~30, 2017.

\bibitem[Liu et~al.(2020)Liu, Zhu, and Belkin]{ntk_linearity}
Liu, C., Zhu, L., and Belkin, M.
\newblock On the linearity of large non-linear models: {W}hen and why the tangent kernel is constant.
\newblock In \emph{Advances in Neural Information Processing Systems}, volume~33, pp.\  15954--15964, 2020.

\bibitem[Liu et~al.(2022)Liu, Chen, and Zhang]{dfl_consensus_2}
Liu, W., Chen, L., and Zhang, W.
\newblock Decentralized federated learning: Balancing communication and computing costs.
\newblock \emph{IEEE Transactions on Signal and Information Processing over Networks}, 8:\penalty0 131--143, 2022.

\bibitem[Martínez~Beltrán et~al.(2023)Martínez~Beltrán, Pérez, Sánchez, Bernal, Bovet, Pérez, Pérez, and Celdrán]{DFL_survey}
Martínez~Beltrán, E.~T., Pérez, M.~Q., Sánchez, P. M.~S., Bernal, S.~L., Bovet, G., Pérez, M.~G., Pérez, G.~M., and Celdrán, A.~H.
\newblock Decentralized federated learning: Fundamentals, state of the art, frameworks, trends, and challenges.
\newblock \emph{IEEE Communications Surveys; Tutorials}, 25\penalty0 (4):\penalty0 2983–3013, 2023.

\bibitem[McMahan et~al.(2017)McMahan, Moore, Ramage, Hampson, and y~Arcas]{fedavg}
McMahan, B., Moore, E., Ramage, D., Hampson, S., and y~Arcas, B.~A.
\newblock Communication-efficient learning of deep networks from decentralized data.
\newblock In \emph{Artificial Intelligence and Statistics}, pp.\  1273--1282. PMLR, 2017.

\bibitem[Mothukuri et~al.(2021)Mothukuri, Parizi, Pouriyeh, Huang, Dehghantanha, and Srivastava]{fl_survey_mothukuri}
Mothukuri, V., Parizi, R.~M., Pouriyeh, S., Huang, Y., Dehghantanha, A., and Srivastava, G.
\newblock A survey on security and privacy of federated learning.
\newblock \emph{Future Generation Computer Systems}, 115:\penalty0 619--640, 2021.

\bibitem[Sattler et~al.(2020)Sattler, Wiedemann, Müller, and Samek]{comm_efficient}
Sattler, F., Wiedemann, S., Müller, K.-R., and Samek, W.
\newblock Robust and communication-efficient federated learning from non-i.i.d. data.
\newblock \emph{IEEE Transactions on Neural Networks and Learning Systems}, 31\penalty0 (9):\penalty0 3400--3413, 2020.

\bibitem[Savazzi et~al.(2020)Savazzi, Nicoli, and Rampa]{dfl_consensus_1}
Savazzi, S., Nicoli, M., and Rampa, V.
\newblock Federated learning with cooperating devices: A consensus approach for massive {IoT} networks.
\newblock \emph{IEEE Internet of Things Journal}, 7\penalty0 (5):\penalty0 4641--4654, 2020.

\bibitem[Shi et~al.(2023)Shi, Shen, Wei, Sun, Yuan, Wang, and Tao]{dfedsam}
Shi, Y., Shen, L., Wei, K., Sun, Y., Yuan, B., Wang, X., and Tao, D.
\newblock Improving the model consistency of decentralized federated learning.
\newblock In \emph{International Conference on Machine Learning}, 2023.

\bibitem[Shiri et~al.(2022)Shiri, Vafaei~Sadr, Amini, Salimi, Sanaat, Akhavanallaf, Razeghi, Ferdowsi, Saberi, Arabi, Becker, Voloshynovskiy, z, Rahmim, and Zaidi]{tumor_example}
Shiri, I., Vafaei~Sadr, A., Amini, M., Salimi, Y., Sanaat, A., Akhavanallaf, A., Razeghi, B., Ferdowsi, S., Saberi, A., Arabi, H., Becker, M., Voloshynovskiy, S., z, D., Rahmim, A., and Zaidi, H.
\newblock {{D}ecentralized distributed multi-institutional {P}{E}{T} image segmentation using a federated deep learning framework}.
\newblock \emph{Clin Nucl Med}, 47\penalty0 (7):\penalty0 606--617, Jul 2022.

\bibitem[Sun et~al.(2021)Sun, Li, and Wang]{dfedavg}
Sun, T., Li, D., and Wang, B.
\newblock Decentralized federated averaging.
\newblock \emph{IEEE Transactions on Pattern Analysis and Machine Intelligence}, 45\penalty0 (4):\penalty0 4289--4301, 2021.

\bibitem[Tan et~al.(2023)Tan, Yu, Cui, and Yang]{pfl_survey}
Tan, A.~Z., Yu, H., Cui, L., and Yang, Q.
\newblock Towards personalized federated learning.
\newblock \emph{IEEE Transactions on Neural Networks and Learning Systems}, 34\penalty0 (12):\penalty0 9587--9603, 2023.

\bibitem[Tirer et~al.(2022)Tirer, Bruna, and Giryes]{ntkresnet}
Tirer, T., Bruna, J., and Giryes, R.
\newblock Kernel-based smoothness analysis of residual networks.
\newblock In \emph{2nd Mathematical and Scientific Machine Learning Conference}, 2022.

\bibitem[Vaswani et~al.(2017)Vaswani, Shazeer, Parmar, Uszkoreit, Jones, Gomez, Kaiser, and Polosukhin]{attention_is_all_you_need}
Vaswani, A., Shazeer, N., Parmar, N., Uszkoreit, J., Jones, L., Gomez, A.~N., Kaiser, L.~u., and Polosukhin, I.
\newblock Attention is all you need.
\newblock In \emph{Advances in Neural Information Processing Systems}, volume~30, 2017.

\bibitem[Xiao et~al.(2017)Xiao, Rasul, and Vollgraf]{fashionmnist}
Xiao, H., Rasul, K., and Vollgraf, R.
\newblock Fashion-{MNIST}: A novel image dataset for benchmarking machine learning algorithms, 2017.

\bibitem[Yang(2019)]{ntk_transformer}
Yang, G.
\newblock Wide feedforward or recurrent neural networks of any architecture are {G}aussian processes.
\newblock In \emph{Advances in Neural Information Processing Systems}, volume~32, 2019.

\bibitem[Yu et~al.(2022)Yu, Wei, Karimireddy, Ma, and Jordan]{yu2022tct}
Yu, Y., Wei, A., Karimireddy, S.~P., Ma, Y., and Jordan, M.
\newblock {TCT}: Convexifying federated learning using bootstrapped neural tangent kernels.
\newblock In \emph{Advances in Neural Information Processing Systems}, 2022.

\bibitem[Yuan et~al.(2024)Yuan, Wang, Sun, Yu, and Brinton]{dfl_survey_2}
Yuan, L., Wang, Z., Sun, L., Yu, P.~S., and Brinton, C.~G.
\newblock Decentralized federated learning: A survey and perspective.
\newblock \emph{IEEE Internet of Things Journal}, 2024.

\bibitem[Yue et~al.(2022)Yue, Jin, Pilgrim, Wong, Baron, and Dai]{yue2022neuraltangentkernelempowered}
Yue, K., Jin, R., Pilgrim, R., Wong, C.-W., Baron, D., and Dai, H.
\newblock Neural tangent kernel empowered federated learning.
\newblock In \emph{International Conference on Machine Learning}, 2022.

\bibitem[Zhang et~al.(2021)Zhang, Xie, Bai, Yu, Li, and Gao]{fl_survey_zhang}
Zhang, C., Xie, Y., Bai, H., Yu, B., Li, W., and Gao, Y.
\newblock A survey on federated learning.
\newblock \emph{Knowledge-Based Systems}, pp.\  106775, 2021.

\bibitem[Zhu et~al.(2019)Zhu, Liu, and Han]{deep_leakage}
Zhu, L., Liu, Z., and Han, S.
\newblock Deep leakage from gradients.
\newblock In \emph{Advances in Neural Information Processing Systems}, volume~32, 2019.

\end{thebibliography}
